\documentclass[]{tencent_hunyuan}

\usepackage[utf8]{inputenc}
\usepackage{amsmath}
\usepackage{amsfonts}
\usepackage{amssymb}
\usepackage{booktabs}
\usepackage{tabularx}
\usepackage{multirow}
\usepackage{enumitem}
\usepackage{float}
\usepackage{algorithm}
\usepackage{algorithmic}
\usepackage[most]{tcolorbox}
\usepackage{advdate}

\newtcolorbox{placeholderbox}{
  enhanced,
  colback=hunyuansky!4!white,
  colframe=hunyuanblue!70!white,
  boxrule=0.5pt,
  arc=2.5mm,
  left=2mm,
  right=2mm,
  top=1.5mm,
  bottom=1.5mm
}

\newcommand{\worldclawfont}{\fontsize{23}{26}\selectfont\bfseries\sffamily}
\newcommand{\worldclawwordmark}{%
  {\worldclawfont\color{hunyuanblue} WorldClaw}%
}
\title{%
  {\worldclawfont\raisebox{-0.28em}{\includegraphics[height=1.4em]{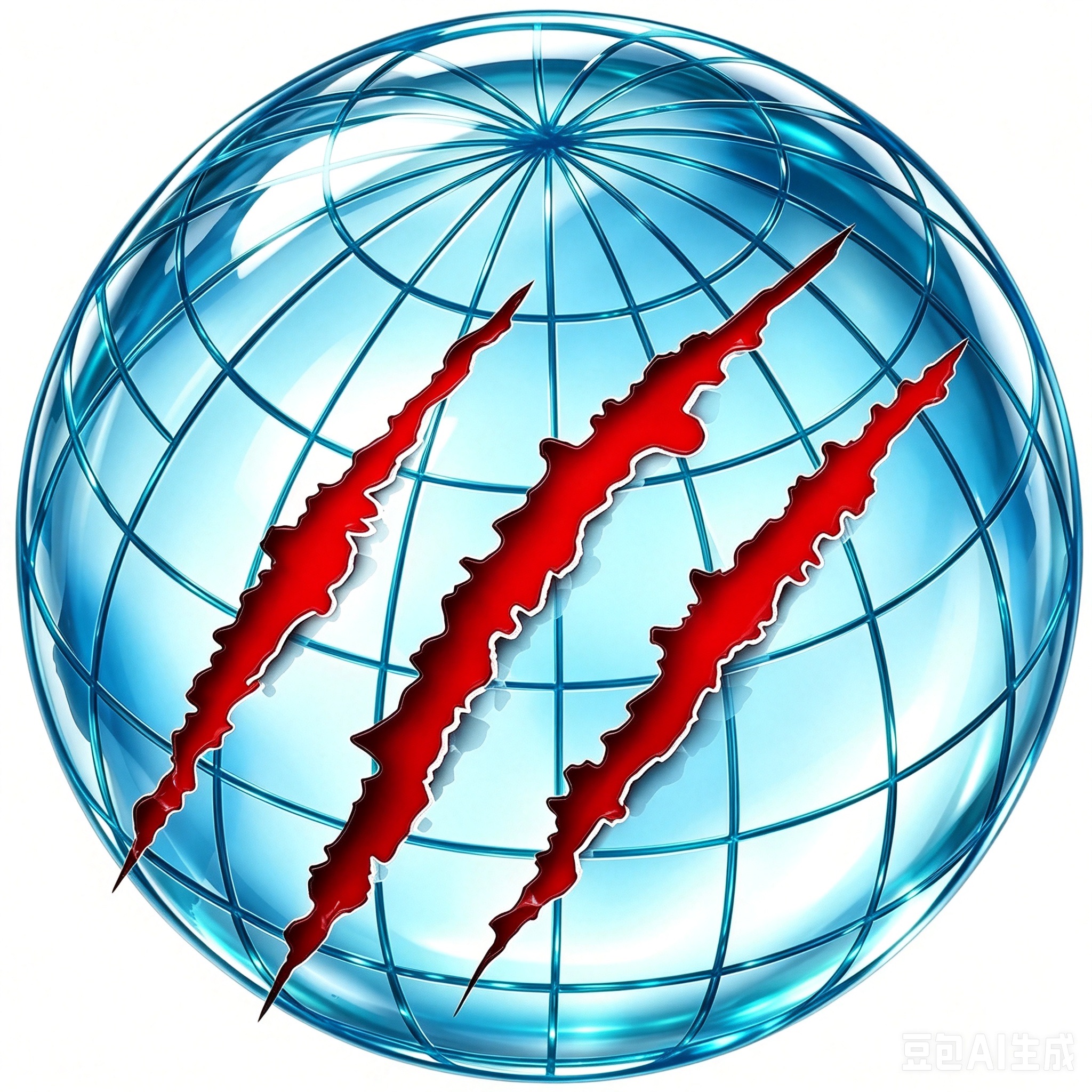}}}%
  \hspace{0.18em}%
  \worldclawwordmark%
  \\[0.45em]%
  Agentic 3D Open-World Generation at Scale%
}
\affiliation{Tencent Hunyuan}

\abstract{
Generating large-scale, freely explorable 3D worlds from open-ended text remains challenging because a system must jointly maintain global spatial coherence, rich local content, and explicit assets suitable for downstream editing and reuse. We present \textbf{WorldClaw}, a fully agentic, coarse-to-fine framework for open-world 3D scene generation. Planning agents translate a text prompt into a structured specification of regions, terrain, assets, materials, and spatial relations. WorldClaw then builds a globally coherent terrain foundation from semantic layouts, reusable assets, generative or procedural materials, and a region-aware height field. For detail-demanding regions, it generates terrain-conditioned compositions, reconstructs editable textured meshes, and recovers their placement on the terrain; render-based agents further refine terrain, objects, appearance, and contacts. Across diverse open-world prompts, WorldClaw produces large-scale scenes with coherent spatial organization, visually compelling local content, and editable instance-level assets while preserving a consistent global terrain structure.
}

\date{\DayAfter[1]}
\checkdata[Project Page]{\url{https://tencent-hunyuan.github.io/Hunyuan3D-WorldClaw}}

\begin{document}
\maketitle

\clearpage
\tableofcontents
\clearpage

\section{Introduction}

Generating freely explorable 3D worlds from a single sentence remains one of the most compelling yet difficult goals in generative content creation. 
Video games, film production, virtual reality, embodied intelligence, and robot simulation all rely on explicit 3D scenes that preserve coherent geometry, appearance, and semantics throughout a continuous space. Unlike an image or a video locked to a predefined camera path, such a world must not only look plausible, but also be walkable, structurally consistent, and editable. 
Together, these properties turn generated content from a fixed visual artifact into a persistent, reusable environment that supports exploration, interaction, simulation, and iterative creation.

Open-world 3D scene generation has been studied extensively for decades and has recently gained renewed attention with the emergence of \textit{3D world models}, which aim to create explorable, immersive visual experiences. Nevertheless, generating large-scale 3D scenes with rich, diverse, and high-quality content remains a challenge. We categorize the existing methods into four categories: Procedural Content Generation (PCG), Image/Video-Lifting, Native 3D Diffusion, and Multi-Modal LLM Agents (MLLM-Agents) based methods.
\textbf{PCG} builds terrain, vegetation, and buildings with rules and executable programs, offering strong controllability and scalability as seen in Infinigen~\cite{raistrick2023infinite}. The PCG methods are often constrained by limited representation capacity, as they rely on hard-coded rules and programs to generate content, which is limited by the expressiveness of the rules and programs. 
\textbf{Image/Video-Lifting} methods leverage image or video generators to synthesize scene observations and lift them into 3D representations such as meshes or 3D Gaussians via depth estimation or multi-view reconstruction. 
Notable products include Marble~\cite{worldlabs2026marble}. Video-generation models offer rich content but often lack global consistency and geometric fidelity. They are also computationally inefficient: for example, generating a complete 3D house scene requires a long video clip that circles the house to capture multiple views, resulting in the generation of many frames. Otherwise, novel viewpoints not covered by the generated video tend to exhibit blurring and incoherent geometry.
\textbf{Native 3D Diffusion} methods learn scene distributions directly over 3D voxels, distance fields, or Gaussians, improving geometric fidelity and cross-view consistency~\cite{ren2024xcube,wu2024blockfusion,meng2025lt3sd}. 
They can efficiently generate high-quality geometry and textures in a single diffusion rollout, but their content diversity is limited by the scarcity of large-scale 3D scene datasets.
\textbf{MLLM Agents} leverage the world knowledge of multimodal large language models for intent understanding, spatial planning, tool orchestration, and render-based refinement, enabling natural-language instructions to drive executable scene-construction pipelines~\cite{yang2024holodeck,yang2026sceneweaver,xia2026sage,pfaff2026scenesmith}. However, despite excelling at global planning and semantic-level task execution, they lack precise 3D spatial control. For example, when asked to adjust a single object’s position, they may overcorrect because of inadequate spatial scale awareness, limiting their ability to generate scenes with complex object relationships. 

In this report, we present \textbf{WorldClaw}, a coarse-to-fine agentic framework for generating explorable 3D worlds from open-ended text. Its design is guided by a central insight: a globally coherent world need not be generated everywhere at once. Shared constraints, including scene semantics, spatial organization, and the terrain foundation, can first be established globally, while the objects and local relationships that distinguish individual  regions are progressively realized where needed. Following this global-to-regional principle, WorldClaw proceeds through three stages: \emph{intent analysis and planning} (Section~\ref{sec:intent-analysis-and-planning}), \emph{global terrain generation} (Section~\ref{sec:global-terrain-generation}), and \emph{regional object generation and placement} (Section~\ref{sec:regional-object-generation-and-placement}). It first transforms the user prompt into a structured scene specification describing regions, terrain, objects, appearance, and spatial relationships; it then constructs an irregular terrain with explicit regional semantics to anchor scene scale and spatial layout; and finally, it generates, places, and refines instance-level content in regions that require further development. Throughout this process, specialized agents translate high-level intent into executable plans, orchestrate the required procedural and generative tools, and inspect and refine the resulting terrain and objects.

At the global terrain generation stage, guided by a structured terrain plan and semantic layout map, WorldClaw combines region-aware procedural terrain construction with generated materials and reusable assets to create irregular, regionally organized landforms. Then Render-based refinement further improves geometry and appearance, yielding a controllable global foundation for subsequent regional content generation.

After establishing the global terrain, WorldClaw selectively populates regions that require instance-level content. For each selected region, it renders the local terrain as a 2D image and uses an image-editing model to populate the image with objects. The editing prompt incorporates both global scene context and local terrain information. The inserted objects are then reconstructed as 3D assets using image-to-3D models and placed on the terrain. This approach enriches the scene with spatially coherent content while preserving its global structure. A subsequent agentic refinement stage further improves object quality and resolves object--terrain contact issues.

The main contributions of this work are summarized as follows:
\begin{itemize}[leftmargin=*,nosep]

    \item We propose a semantic-layout-guided procedural terrain generator that creates controllable, region-aware landforms across spatial scales.

    \item We translate global scene and local terrain constraints into executable regional plans that yield diverse, spatially coherent object layouts with fine-grained regional control.

    \item We introduce an agentic refinement loop that iteratively improves terrain and object quality, corrects object scale and pose, and resolves object-terrain contact issues.

\end{itemize}

WorldClaw represents scenes as independently editable textured meshes with explicit terrain placements, supporting free-viewpoint rendering, asset reuse, and conventional game-engine workflows.

\section{Method}

Given an open-ended user prompt \(q\), WorldClaw aims to construct an explicit 3D world \(\mathcal{S}\) whose global organization, terrain geometry, surface appearance, and object composition are consistent with the user intent. The resulting world should support free-viewpoint exploration and preserve terrain and object instances as editable 3D content. Rather than synthesizing the entire world in a single pass, WorldClaw adopts a coarse-to-fine, global-to-regional construction strategy. It first establishes the scene-wide semantics, regional organization, and terrain foundation, and then selectively realizes detailed instance-level content in regions that require further development.

\begin{figure}[H]
    \centering
    \includegraphics[width=\linewidth]{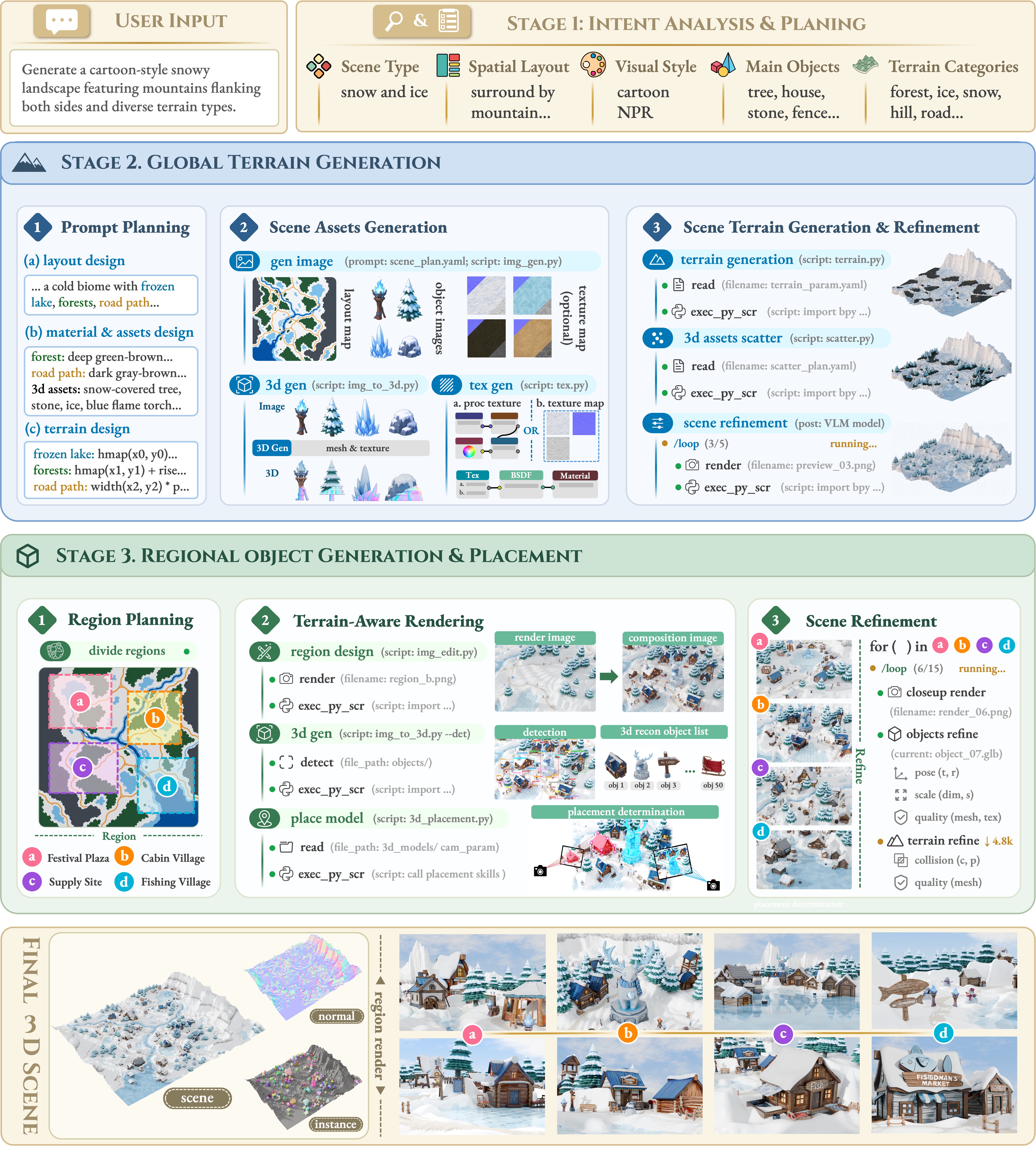}
    \caption{\textbf{Overview of WorldClaw.}
    Given an open-ended text prompt, WorldClaw constructs an explicit, explorable, and editable 3D world through a three-stage global-to-regional pipeline.
    (1) \textit{Intent Analysis and Planning} translates the prompt into a structured scene specification.
    (2) \textit{Global Terrain Generation} establishes a region-aware terrain foundation with coherent geometry, appearance, and spatial semantics.
    (3) \textit{Regional Object Generation and Placement} populates planned regions with editable 3D assets and refines their arrangements and interactions with terrain.}
    \label{fig:pipeline}
\end{figure}

As illustrated in Fig.~\ref{fig:pipeline}, the construction process consists of three sequential stages:
\begin{equation}
    \mathcal{P}
    = F_{\mathrm{plan}}(q), \qquad
    \mathcal{T}
    = F_{\mathrm{terrain}}(\mathcal{P}), \qquad
    \mathcal{O}
    = F_{\mathrm{region}}(\mathcal{P}, \mathcal{T}),
\end{equation}
where \(\mathcal{P}\) denotes the structured scene specification, \(\mathcal{T}\) denotes the global terrain representation, and \(\mathcal{O}\) denotes the set of generated and placed regional object instances. The final world is composed as
\begin{equation}
    \mathcal{S} = \operatorname{Compose}(\mathcal{T}, \mathcal{O}).
\end{equation}

The \textit{intent analysis and planning} stage (Section~\ref{sec:intent-analysis-and-planning}) converts \(q\) into \(\mathcal{P}\), which specifies the regions, terrain requirements, object categories, appearance attributes, and spatial relationships of the target world. The \textit{global terrain generation} stage (Section~\ref{sec:global-terrain-generation}) instantiates the terrain and regional constraints in \(\mathcal{P}\) as a coherent terrain foundation \(\mathcal{T}\). Conditioned on both \(\mathcal{P}\) and \(\mathcal{T}\), the \textit{regional object generation and placement} stage (Section~\ref{sec:regional-object-generation-and-placement}) constructs the object set \(\mathcal{O}\), determines instance-level placements, and refines local object--terrain relationships. Specialized agents execute the three stages and communicate through shared structured intermediate representations, allowing local generation and refinement to preserve the globally established scene organization.

\subsection{Intent Analysis and Planning}
\label{sec:intent-analysis-and-planning}

Users typically describe a target scene using a short, open-ended text prompt, whereas constructing a complete 3D outdoor scene requires substantially more detailed specifications, including terrain structure, regional organization, object composition, material style, environmental atmosphere, and spatial relationships. Directly passing the raw prompt to downstream generation modules introduces two major challenges. First, users often specify only a small number of salient concepts while omitting the spatial, geometric, and appearance attributes required for scene construction. Second, category references, qualitative spatial relationships, and high-level style descriptions may be ambiguous and interpreted inconsistently across different generation stages. Inspired by prior work~\cite{huang2026majutsucity,he2026mind,feng2026gen,ye2026genclaw}, WorldClaw introduces an intent analysis and planning module 
composed of an intent analysis agent and a scene planning agent. Together, they convert the open-ended user prompt into a structured scene specification \(\mathcal{P}\), which serves as the shared semantic interface for subsequent terrain construction and regional object generation.

\paragraph{\textbf{Intent Analysis.}}
Given the user prompt \(q\), the intent analysis agent extracts and normalizes the constraints explicitly expressed by the user, including the scene type, theme and visual style, key regions and objects, spatial relationships, and user preferences. This stage only summarizes information already contained in the prompt and neither introduces new scene content nor completes unspecified attributes. By separating explicit constraint extraction from missing-information completion, it preserves the original user intent and provides reliable constraints for subsequent scene planning.

\paragraph{\textbf{Scene-Level Planning.}}
Conditioned on the original prompt and its explicit constraints, the scene planning agent resolves ambiguous descriptions while preserving the user requirements and completes unspecified information required by downstream generation modules according to a predefined scene-specification schema. The resulting scene plan is organized as
\begin{equation}
\mathcal{P}
=
\left(
\mathcal{R},
\mathcal{C}_{\mathrm{terrain}},
\mathcal{C}_{\mathrm{object}}
\right),
\label{eq:scene-level-plan}
\end{equation}

where \(\mathcal{R}\) describes the major scene regions, their attributes, and their spatial relationships. The terrain specification \(\mathcal{C}_{\mathrm{terrain}}\) describes the terrain types, landform characteristics, surface appearance, and terrain-associated assets required by each region, while the object specification \(\mathcal{C}_{\mathrm{object}}\) specifies the required object categories, appearance attributes, approximate densities, and region-level spatial relationships. Global properties, including the scene theme, overall visual style, material preferences, and environmental atmosphere, are retained as shared attributes of the corresponding fields, ensuring that terrain and object generation follow consistent visual and semantic conditions.

This stage produces the structured scene specification required by subsequent scene construction, rather than directly generating terrain geometry or object instances. The resulting specification \(\mathcal{P}\) serves as a shared set of semantic and spatial constraints, ensuring a consistent interpretation of the user intent, regional organization, visual style, and scene content across downstream modules.
\subsection{Global Terrain Generation}
\label{sec:global-terrain-generation}

Terrain is not only the geometric foundation of a 3D world, but also a global structure that organizes regional semantics, spatial hierarchy, surface appearance, and environmental content. Large-scale landforms determine the overall world shape, perceived scale, and explorable extent, while local relief and surface details further affect navigability, visual realism, and stylistic expression. However, many existing scene-generation pipelines simplify the ground as a planar or weakly varying support and organize scene content primarily by placing objects on top of it. Such planar assumptions cannot adequately represent geometrically expressive landforms, including mountains, canyons, dunes, and terraces, and limit the continuous yet semantically distinct organization of different regions.

WorldClaw therefore introduces a global terrain generation module that transforms the structured scene specification \(\mathcal{P}\) produced by intent analysis and planning into a global terrain representation \(\mathcal{T}\). The module progressively realizes the terrain requirements through explicit, inspectable, and editable intermediate representations. It comprises three stages: terrain planning refines high-level semantic constraints into an executable terrain specification; terrain asset generation constructs a semantic layout map, reusable terrain-asset prototypes, and surface materials; and terrain generation and refinement instantiates these conditions as a renderable 3D terrain with region-aware semantics, composite landforms, surface appearance, and environmental details. The resulting \(\mathcal{T}\) provides a shared geometric foundation and spatial semantics for subsequent regional content generation in Section~\ref{sec:regional-object-generation-and-placement}.

\subsubsection{Terrain Planning}
\label{sec:terrain-planning}
The structured scene specification \(\mathcal{P}\) describes the regional composition, spatial relationships, and high-level terrain requirements of the target world, but does not yet provide all information required for direct terrain construction. For example, it may not fully specify region boundaries, world scale, landform parameters, asset densities, or material implementations. Directly generating terrain from such high-level constraints would require a single generative process to simultaneously perform semantic interpretation, missing-information completion, and spatial and geometric instantiation. Since the same semantic description may admit multiple plausible terrain designs, the result can appear visually reasonable while deviating from the desired regional relationships, landform hierarchy, or content composition. To bridge the gap between high-level scene semantics and executable terrain construction, we employ a terrain planning agent to produce a structured terrain specification.

\paragraph{\textbf{Structured Terrain Specification.}}
The terrain planning agent converts the terrain-related constraints in \(\mathcal{P}\) into a structured terrain specification
\begin{equation}
\mathcal{P}_{\mathrm{terrain}}
=
\left(
\mathbf{p}_{\mathrm{layout}},
\mathbf{p}_{\mathrm{asset}},
\mathbf{p}_{\mathrm{material}},
\boldsymbol{\theta}_{\mathrm{terrain}}
\right),
\label{eq:terrain-specification}
\end{equation}
where \(\mathbf{p}_{\mathrm{layout}}\) specifies region categories, relative positions, adjacency relationships, and approximate coverage; \(\mathbf{p}_{\mathrm{asset}}\) describes terrain-associated asset categories, regional affinities, and target densities; \(\mathbf{p}_{\mathrm{material}}\) defines the surface types, visual styles, and texture requirements of different regions; and \(\boldsymbol{\theta}_{\mathrm{terrain}}\) contains numerical parameters such as world scale, region-specific base elevations, noise frequencies and amplitudes, geomorphic operators and their weights, and boundary-blending widths. We define a standardized schema for these components so that \(\mathcal{P}_{\mathrm{terrain}}\) serves as an explicit interface among terrain planning, asset generation, and geometric construction.

\paragraph{\textbf{Tool-Augmented Planning.}}
When the scene specification contains concepts that require external knowledge, the agent invokes a search tool to retrieve relevant references. When textual constraints are insufficient to express complex spatial relationships or visual styles, the agent additionally generates a scene concept image \(\mathbf{I}_{\mathrm{concept}}\) as visual conditioning for landform composition, regional layout, and artistic style. The retrieved references and concept image supplement \(\mathcal{P}_{\mathrm{terrain}}\) without replacing explicit user constraints; \(\mathbf{I}_{\mathrm{concept}}\) is subsequently used as an optional condition for terrain asset generation and regional planning.

\subsubsection{Terrain Asset Generation}
\label{sec:terrain-asset-generation}
To convert the textual terrain specification into explicit visual and geometric conditions, the terrain asset generation agent takes \(\mathcal{P}_{\mathrm{terrain}}\) and the optional concept image \(\mathbf{I}_{\mathrm{concept}}\) as conditions and produces
\begin{equation}
\mathcal{A}_{\mathrm{terrain}}
=
\left(
    \mathbf{I}_{\mathrm{layout}},
    \mathcal{I}_{\mathrm{asset}},
    \mathcal{O}_{\mathrm{asset}},
    \mathcal{M}_{\mathrm{terrain}}
\right),
\label{eq:terrain-asset-generation}
\end{equation}
where \(\mathbf{I}_{\mathrm{layout}}\) is a scene layout map that constrains regional distributions, \(\mathcal{I}_{\mathrm{asset}}\) is a set of representative asset images, \(\mathcal{O}_{\mathrm{asset}}\) is a set of 3D assets intended for subsequent global scattering, and \(\mathcal{M}_{\mathrm{terrain}}\) is a set of materials defining surface appearance. These intermediate assets are constructed using image, 3D, and material generation capabilities and serve as explicit inputs to subsequent terrain construction. Both \(\mathbf{I}_{\mathrm{layout}}\) and \(\mathcal{I}_{\mathrm{asset}}\) are generated by GPT-Image-2~\cite{openai2026gptimage2}

\paragraph{\textbf{Scene Layout Map.}}
Indoor-scene generation commonly represents spatial layouts using floor plans or structured room representations \cite{xia2026sage,pfaff2026scenesmith}. These representations are designed primarily for regular walls and approximately planar regions and are therefore not directly suited to natural terrain with curved boundaries and irregular shapes. We instead generate a global semantic layout map \(\mathbf{I}_{\mathrm{layout}}\) from \(\mathbf{p}_{\mathrm{layout}}\), using distinct colors to encode predefined terrain categories. The layout map converts textual descriptions of region categories, relative positions, adjacency relationships, and coverage into a unified 2D spatial partition, which is subsequently used for region-mask extraction, height-field generation, material assignment, and asset scattering.

\paragraph{\textbf{Terrain Asset Prototypes.}}
For environmental elements that are repeatedly instantiated across the global terrain, this stage generates reusable 3D asset prototypes without determining instance-specific positions, scales, or orientations. Given the asset requirements \(\mathbf{p}_{\mathrm{asset}}\), the agent first generates a representative image set \(\mathcal{I}_{\mathrm{asset}}\), including terrain-associated categories such as rocks, vegetation clusters, and landform attachments. The system then uses the image-to-3D capability of Hunyuan3D\cite{hunyuan3d2025hunyuan3d,lai2025hunyuan3d} to convert these references into the reusable 3D asset-prototype set \(\mathcal{O}_{\mathrm{asset}}\), which provides geometric candidates for subsequent environmental scattering.

\paragraph{\textbf{Terrain Materials.}}
To balance expressive local surfaces with scalable large-area coverage, we employ two complementary pathways, namely generative texture synthesis and procedural material generation, to construct \(\mathcal{M}_{\mathrm{terrain}}\) according to \(\mathbf{p}_{\mathrm{material}}\). The generative pathway produces texture channels such as albedo, normal, and roughness maps for local surfaces with complex appearance or irregular details. The procedural pathway programmatically assembles Blender material nodes to create tileable and parameter-adjustable surface materials for large-scale regions. Materials from both pathways are associated with the semantic regions in \(\mathbf{I}_{\mathrm{layout}}\), ensuring consistency between regional organization and surface appearance.

\subsubsection{Terrain Generation \& Refinement}
\label{sec:terrain-generation-refinement}

\begin{figure}[ht]
    \centering\includegraphics[width=1\textwidth]{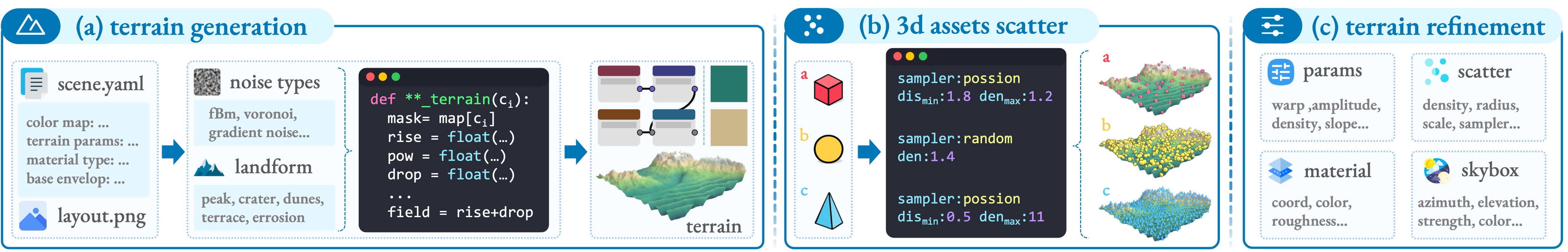}
    \caption{\textbf{Overview of global terrain generation and refinement.}(a) \textit{Initial Height-Field Generation} constructs composite terrain geometry from the semantic layout map and region-specific terrain parameters and assigns materials to the corresponding regions.(b) \textit{Global Terrain Asset Scattering} instantiates terrain-associated assets according to regional semantics and local surface conditions.(c) \textit{Terrain Refinement} renders, inspects, and locally edits the terrain to correct geometric transitions, material scales, asset distributions, and rendering artifacts.}
    \label{fig:global-terrain-generation-and-refinement}
\end{figure}

Given the structured terrain specification \(\mathcal{P}_{\mathrm{terrain}}\) and the intermediate asset package \(\mathcal{A}_{\mathrm{terrain}}\), the terrain generation agent constructs the global terrain representation \(\mathcal{T}\). The representation comprises terrain geometry, regional semantics, surface materials, and scattered terrain-associated assets. The process consists of initial height-field generation, global terrain asset scattering, and terrain refinement.

\paragraph{\textbf{Initial Height-Field Generation.}}
As illustrated in Fig.~\ref{fig:global-terrain-generation-and-refinement}(a), the agent parses the scene layout map \(\mathbf{I}_{\mathrm{layout}}\), maps its color codes to predefined terrain categories, and extracts a mask for each region \(r\). Boundary smoothing produces normalized soft weights \(\widetilde{m}_{r}\), allowing height-field generation, material assignment, and subsequent asset scattering to share the same semantic partition. Given the region-level terrain parameters \(\boldsymbol{\theta}_{\mathrm{terrain}}\), the global height field is represented as
\begin{equation}
H(\mathbf{x})
=
\sum_{r}
\widetilde{m}_{r}(\mathbf{x})
\left[
h_{r}
+
\sum_{k} w_{r,k}N_{r,k}(\mathbf{x})
+
\sum_{j}\alpha_{r,j}G_{r,j}(\mathbf{x})
\right],
\label{eq:region-aware-height-field}
\end{equation}
where \(\mathbf{x}\) denotes a 2D position in the terrain domain, \(h_r\) is the base elevation of region \(r\), \(N_{r,k}\) is a noise component at a particular spatial frequency, \(G_{r,j}\) is a geomorphic operator such as peak, dune, terrace, or erosion, and \(w_{r,k}\) and \(\alpha_{r,j}\) control the respective contributions of noise and geomorphic operators~\cite{Patel-2015}. Region-specific composition allows different landforms to form a continuous composite height field with irregular boundaries, while explicit scale parameters allow the same construction logic to accommodate different spatial extents. The same regional weights are subsequently used to assign and blend materials from \(\mathcal{M}_{\mathrm{terrain}}\) over the corresponding surfaces.

\paragraph{\textbf{Global Terrain Asset Scattering.}}
After establishing the base terrain geometry, the system supplements it with mid-scale environmental details using reusable prototypes from \(\mathcal{O}_{\mathrm{asset}}\), as illustrated in Fig.~\ref{fig:global-terrain-generation-and-refinement}(b). This stage handles only basic elements closely associated with terrain and ecological distributions, such as rocks, vegetation clusters, and landform attachments. Objects with explicit functions, instance identities, or complex spatial relationships are deferred to the subsequent regional object generation stage. For each terrain-asset category, the agent samples candidate locations within the corresponding layout masks according to the regional affinities and target densities in \(\mathbf{p}_{\mathrm{asset}}\). Local elevation, slope, and surface normals are then used to filter or adjust the sampled instances. Their final scales and orientations are adapted to the local surface to reduce floating, penetration, and unnatural distributions.

\paragraph{\textbf{Terrain Refinement.}}
As illustrated in Fig.~\ref{fig:global-terrain-generation-and-refinement}(c), the initial terrain may still contain abrupt regional transitions, inconsistent landform scales, mismatched texture proportions, unnatural asset distributions, or local rendering artifacts. Powered by BlenderMCP~\cite{ahujasid2025blendermcp}, the terrain refinement agent re-renders the scene from predefined viewpoints, inspects its geometry, materials, scattering results, and rendering configuration, and performs localized corrections according to the detected issues. Editable variables include region-specific terrain parameters, boundary blending, material texture scales, asset densities and transforms, and, when necessary, environment lighting and rendering settings. The loop continues until no substantial issue is detected or a predefined iteration budget is reached, improving local geometry and visual quality while preserving the semantic layout and global spatial organization.
\subsection{Regional Object Generation and Placement}
\label{sec:regional-object-generation-and-placement}

Section~\ref{sec:global-terrain-generation} constructs a global terrain with explicit region-level semantics, but primarily focuses on large-scale landforms, basic surface appearance, and terrain-associated elements, leaving fine-grained scene objects with explicit functions and spatial relationships uninstantiated. To populate selected regions while preserving the global terrain structure, we introduce a regional object generation and placement module. The module combines the scene specification \(\mathcal{P}\) from Section~\ref{sec:intent-analysis-and-planning} with the generated terrain \(\mathcal{T}\), and plans, generates, and edits only those regions that require further development. This regionalized formulation avoids processing the complete scene in a single pass and provides fine-grained control over regional functions, object compositions, spatial layouts, and visual styles. The overall process comprises regional planning, object generation and placement, and scene refinement.

\subsubsection{Regional Planning}

The region set \(\mathcal{R}\) and object requirements \(\mathcal{C}_{\mathrm{object}}\) in the scene specification describe the global regional organization and high-level object requirements, but do not yet determine which regions should be refined or provide executable regional object configurations. The regional planning agent therefore jointly examines \(\mathcal{P}\) and \(\mathcal{T}\), prioritizes regions with uninstantiated object requirements whose local terrain can support the requested functions, selects a subset \(\mathcal{R}^{+}\subseteq\mathcal{R}\) for further development, and refines the scene-level constraints into

\begin{equation}
\mathcal{P}_{\mathrm{regional}}
=
\left\{
\left(
r,
\phi_{r},
\mathcal{C}^{\mathrm{object}}_{r},
\mathbf{p}^{\mathrm{spatial}}_{r},
\mathbf{p}^{\mathrm{appearance}}_{r}
\right)
\;\middle|\;
r\in\mathcal{R}^{+}
\right\},
\label{eq:regional-specification}
\end{equation}

where each tuple, denoted \(\mathcal{P}_r\), corresponds to a selected region \(r\in\mathcal{R}^{+}\): \(\phi_r\) denotes its functional role, \(\mathcal{C}^{\mathrm{object}}_r\) specifies the categories, counts, and densities of objects to be introduced, \(\mathbf{p}^{\mathrm{spatial}}_r\) describes object--object and object--terrain relationships, and \(\mathbf{p}^{\mathrm{appearance}}_r\) defines region-level appearance and style requirements. Regional planning completes only the region-level information that remains uninstantiated in the scene specification, while preserving the existing region semantics and relationships in \(\mathcal{R}\). When textual constraints are insufficient to specify the desired appearance, the optional concept image \(\mathbf{I}_{\mathrm{concept}}\) supplements the regional specification as a visual reference without replacing explicit constraints derived from the user intent. Consequently, \(\mathcal{P}_{\mathrm{regional}}\) serves as an explicit interface between global scene planning and subsequent object generation and placement.

\subsubsection{Object Generation \& Placement}

To support controllable object generation and placement over the generated terrain, we decompose this stage into three sequential and reusable skills: \textit{region composition}, \textit{object generation}, and \textit{object placement}. Region composition produces a terrain-conditioned 2D layout prior, object generation separates and reconstructs individual instances from the composition image, and object placement recovers their positions, scales, and orientations in the terrain coordinate system. For each region \(r\), this stage produces an object set

\begin{equation}
\mathcal{O}_{r}
=
\left\{
\left(
M_i,
\mathcal{U}_i,
T_{\mathrm{place}}^{i}
\right)
\right\}_{i=1}^{n_r},
\label{eq:regional-object-set}
\end{equation}

where \(M_i\), \(\mathcal{U}_i\), and \(T_{\mathrm{place}}^{i}\) denote the geometry, appearance attributes, and placement transformation of object \(i\), respectively. By coordinating the three skills, the regional agent progressively converts \(\mathcal{P}_r\) into a 3D object layout consistent with both the synthesized composition and the underlying terrain, while retaining an instance-level representation for subsequent editing and refinement.

\paragraph{\textbf{Region Composition.}}

For each selected region \(r\), the agent first renders the existing terrain and records the associated camera parameters \(\kappa_r=(K_t,E_t)\), producing a terrain image \(\mathbf{I}^{\mathrm{terrain}}_r\). In contrast to directly synthesizing an isolated regional image, this rendering preserves local topography, material appearance, viewing direction, and surrounding spatial context. The agent then converts \(\mathcal{P}_r\) into a structured regional prompt and generates a region composition image conditioned on the terrain rendering, regional specification, and optional concept image:

\begin{equation}
\mathbf{I}^{\mathrm{comp}}_r
=
\mathcal{G}_{\mathrm{image}}
\left(
\mathbf{I}^{\mathrm{terrain}}_r,
\mathcal{P}_r,
\mathbf{I}_{\mathrm{concept}}
\right).
\label{eq:regional-composition}
\end{equation}

The terrain rendering and camera parameters provide explicit guidance for retaining the underlying topography and viewpoint, while the regional specification controls the categories, styles, densities, and spatial relationships of newly introduced content. The composition \(\mathbf{I}^{\mathrm{comp}}_r\) is not treated as final 3D geometry; instead, it serves as an explicit 2D layout prior that jointly constrains object appearance and spatial organization in the subsequent reconstruction and placement stages.

\paragraph{\textbf{Object Generation.}}

The region composition typically contains multiple objects with substantially different scales and poses. Reconstructing the complete image as a single 3D scene makes it difficult to preserve instance-level reconstruction quality, separate object boundaries, and recover accurate placements. We therefore employ text-guided SAM3~\citep{carion2025sam3} to extract individual 2D object instances from \(\mathbf{I}^{\mathrm{comp}}_r\). In addition to full-image inference, overlapping sliding-window inference improves recall across object scales. Local predictions are mapped back to the region-composition image coordinate system and then deduplicated and merged according to semantic labels and spatial overlap.

For each segmented instance \(i\), we construct an object-centric image \(\mathbf{I}_i\) and mask \(S_i\) by cropping and enlarging its corresponding region. Let \(A_i\) denote the homogeneous affine transformation from the region-composition coordinates to the object-centric image coordinates. Given the regional terrain-camera intrinsics \(K_t\), the equivalent intrinsics after cropping are
\begin{equation}
\widehat{K}_{i}
=
A_iK_t.
\label{eq:focused-camera-intrinsics}
\end{equation}

In homogeneous image coordinates, a pixel in the object-centric image can be mapped back to the region composition using \(A_i^{-1}\). Because cropping and resizing modify only the image coordinate system, the camera extrinsics remain \(E_t\). Recording \(A_i\) and \(\widehat{K}_i\) therefore preserves the original image location and its geometric relationship with the terrain camera when a small object is enlarged.

SAM3D~\citep{chen2026sam3d} then takes \(\mathbf{I}_i\) and \(S_i\) as input and predicts an object mesh \(M_i\), appearance attributes \(\mathcal{U}_i\), a local-to-object-camera transformation \(T_{l2c}^{i}\), and reconstruction-camera intrinsics \(K_i^{o}\). We distinguish the cropped terrain camera from the object reconstruction camera internally estimated by SAM3D. Let \(M_i^{c}=T_{l2c}^{i}M_i\) denote the reconstructed mesh in the object-camera space. Because single-view reconstruction may introduce scale deviations, we perform image-space scale calibration. Let \(\mathcal{B}(\mathbf{I})\) denote the ratio between the foreground bounding-box area and the image area, and let \(b_i^{\mathrm{ref}}=\mathcal{B}(\mathbf{I}_i)\). We iteratively adjust a scale factor \(\lambda_i\) about the mesh center until
\begin{equation}
-\epsilon_i^{-}
\leq
\mathcal{B}\!\left(\Pi\!\left(K_i^{o},\lambda_i M_i^{c}\right)\right)-b_i^{\mathrm{ref}}
\leq
\epsilon_i^{+},
\label{eq:image-space-scale-calibration}
\end{equation}

where \(\Pi(\cdot)\) denotes rendering projection, and \(\epsilon_i^{-}\) and \(\epsilon_i^{+}\) control the tolerances for under- and over-projection, respectively. The asymmetric tolerance more strongly suppresses oversized reconstructions while allowing slight under-coverage caused by segmentation boundaries and single-view ambiguity. This calibration corrects the reconstructed scale within the object-camera coordinate system; the subsequent placement scale transfers the calibrated object from the object-camera coordinate system to the terrain-camera coordinate system.

\paragraph{\textbf{Object Placement.}} After generating each object, we recover its 3D placement using a pair of corresponding rays from the object reconstruction camera and the regional terrain camera. We first cast a ray through the object-center pixel and intersect it with the camera-space mesh \(M_i^c\), obtaining an object reference point \(P_o\) and its depth \(Z_o\). The focused-image center is then mapped back to the region composition using \(A_i^{-1}\), and a second ray is cast from the terrain camera through the mapped pixel. The nearest positive intersection with the terrain mesh \(M_t\) defines the target terrain anchor \(P_t\) and its depth \(Z_t\). Because \(\widehat{K}_i=A_iK_t\) and the extrinsics remain unchanged, the ray from the focused-image center is geometrically equivalent to this inverse-mapped ray. This property allows high-resolution object reconstruction without sacrificing subsequent placement accuracy.

Under the assumptions of approximately square pixels and locally uniform perspective scaling, the initial 3D scale that preserves the apparent object size between the object-camera and terrain-camera coordinate systems is
\begin{equation}
s_i
=
\frac{Z_t}{Z_o}
\frac{f_i^{o}}{\widehat{f}_i},
\label{eq:object-placement-scale}
\end{equation}

where \(f_i^{o}\) denotes the focal length of the object reconstruction camera, while \(\widehat{f}_i\) is obtained directly from the equivalent intrinsics \(\widehat{K}_i\). Let \(R_i\) denote the relative rotation determined by the two camera extrinsics. For a mesh retained in the local coordinate system of SAM3D, the complete placement transformation is
\begin{equation}
T_{\mathrm{place}}^{i}
=
\begin{bmatrix}
s_iR_i & P_t-s_iR_iP_o \\
\mathbf{0}^{\mathsf T} & 1
\end{bmatrix}
T_{l2c}^{i}.
\label{eq:complete-placement-transform}
\end{equation}

The translation term maps \(P_o\) exactly to \(P_t\). If \(T_{l2c}^{i}\) has already been baked into the exported mesh vertices, the rightmost \(T_{l2c}^{i}\) is omitted to avoid a duplicate transformation. Finally, to reduce slight floating caused by mesh discretization and single-view depth errors, we jointly search the anchor depth and isotropic scale along the terrain-camera ray while preserving the 2D projection. The search terminates when the contact ratio between the bottom object voxels and terrain reaches a threshold; otherwise, the candidate with the highest contact ratio is retained.

\subsubsection{Scene Refinement}

\begin{figure}[H]
    \centering
    \includegraphics[width=1\textwidth]{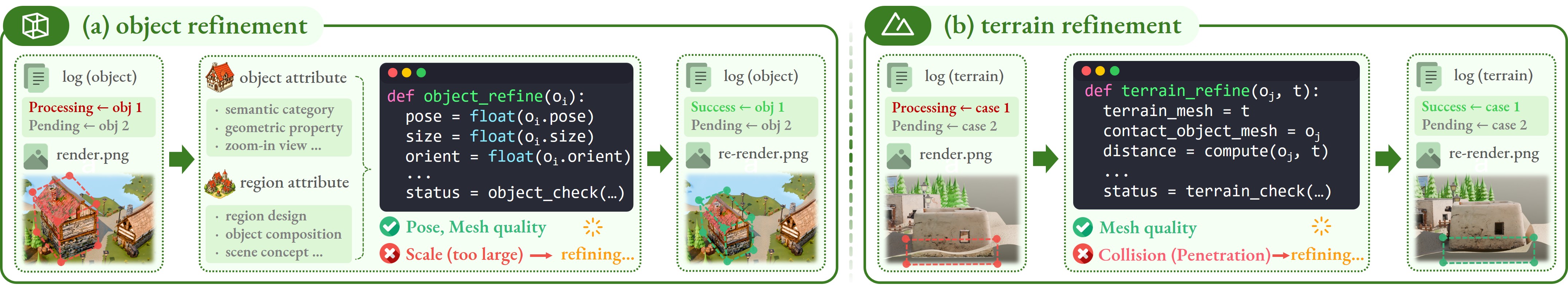}
    \caption{\textbf{Scene refinement.} (a) \textit{Object Refinement} processes objects from a report queue, evaluates pose, mesh quality, and scale against semantic and regional context, applies targeted edits, and verifies the result through re-rendering. (b) \textit{Terrain Refinement} examines support-surface quality and object-terrain collisions, applies local co-deformation to defects such as floating, and updates the report after re-rendering.}
    \label{fig:stage2-3_method}
\end{figure}

Although the preceding stages provide independent object meshes and their initial terrain-aligned placements, a regional scene may still contain inconsistent object scales or poses, insufficient local geometry or appearance quality, and object-terrain contact defects such as floating, excessive penetration, or unstable support. We therefore introduce a scene refinement agent that connects to a 3D engine such as Blender through an executable interface such as the Model Context Protocol (MCP). As shown in Fig.~\ref{fig:stage2-3_method}, the agent maintains a task queue consisting of diagnostic renders and status reports, and performs object refinement followed by terrain refinement.

\paragraph{\textbf{Object Refinement.}}
As illustrated in Fig.~\ref{fig:stage2-3_method}(a), for each pending object, the agent evaluates pose, mesh quality, and scale using its semantic category and geometric properties together with the design and object composition of the surrounding region. Objects with inconsistent scales or implausible orientations are corrected by updating their placement transformations. For SAM3D reconstructions with insufficient geometry or appearance quality, we further condition Hunyuan3D on both the scale-calibrated coarse mesh and the object-centric image. The coarse mesh provides a structural constraint, while the image supplies shape and appearance cues, allowing Hunyuan3D~\cite{hunyuan3d2025hunyuan3d, lai2025hunyuan3d} to improve local surface geometry and generate higher-quality textures or PBR materials while limiting unintended changes to object identity and overall shape. The refined asset inherits the existing \(T_{\mathrm{place}}^{i}\), and can therefore replace the coarse reconstruction without repeating placement. The agent then re-renders the processing object and updates its report until pose, mesh quality, and scale pass the corresponding checks.

\paragraph{\textbf{Terrain Refinement.}}
Once all objects in the pending queue have been refined, the agent proceeds to terrain refinement. As shown in Fig.~\ref{fig:stage2-3_method}(b), this stage evaluates terrain-surface quality together with object--terrain collision and contact, focusing on floating, excessive penetration, and unstable support. When a defect is detected, the agent performs object--terrain co-deformation within the local support region. Depending on the object category and contact state, the object may be vertically repositioned or partially embedded, while the supporting terrain is locally displaced, flattened, or smoothed to conform to the object footprint. All modifications are restricted to the support region to preserve the global landform and neighboring terrain structures. After each edit, the agent re-renders the scene from diagnostic viewpoints and updates the terrain report according to the support-surface and collision checks. Refinement continues until the current task passes the checks or reaches the predefined iteration budget.

\section{Experiments}
\label{sec:experiments}

In this section, we first report the implementation details of WorldClaw, including the foundation models, scene-construction configuration, refinement settings and rendering setup. We then present qualitative results across diverse open-world prompts from both global orbit and local walk views. Finally, we compare WorldClaw with representative text-driven 3D scene generation methods.

\subsection{Implementation Details}

WorldClaw uses Claude Opus 4.8~\cite{anthropic2026opus48} as the underlying agent model. We develop a set of task-specific agent skills that extend the agent with the pretrained foundation models (GPT-Image-2, SAM3, SAM3D, and Hunyuan3D) and executable 3D tools required for scene generation. During object quality refinement based on HunYuan3D, we provide \(2048\times2048\) PBR texture maps for large objects and \(1024\times1024\) maps for small objects. All experiments are run on a server equipped with 4 NVIDIA H20 GPUs. Terrain generation, object generation and placement, scene refinement and image rendering are conducted in Blender 5.1.1.

\subsection{Qualitative Results}

\begin{figure}[H]
    \centering
    \includegraphics[width=1\linewidth]{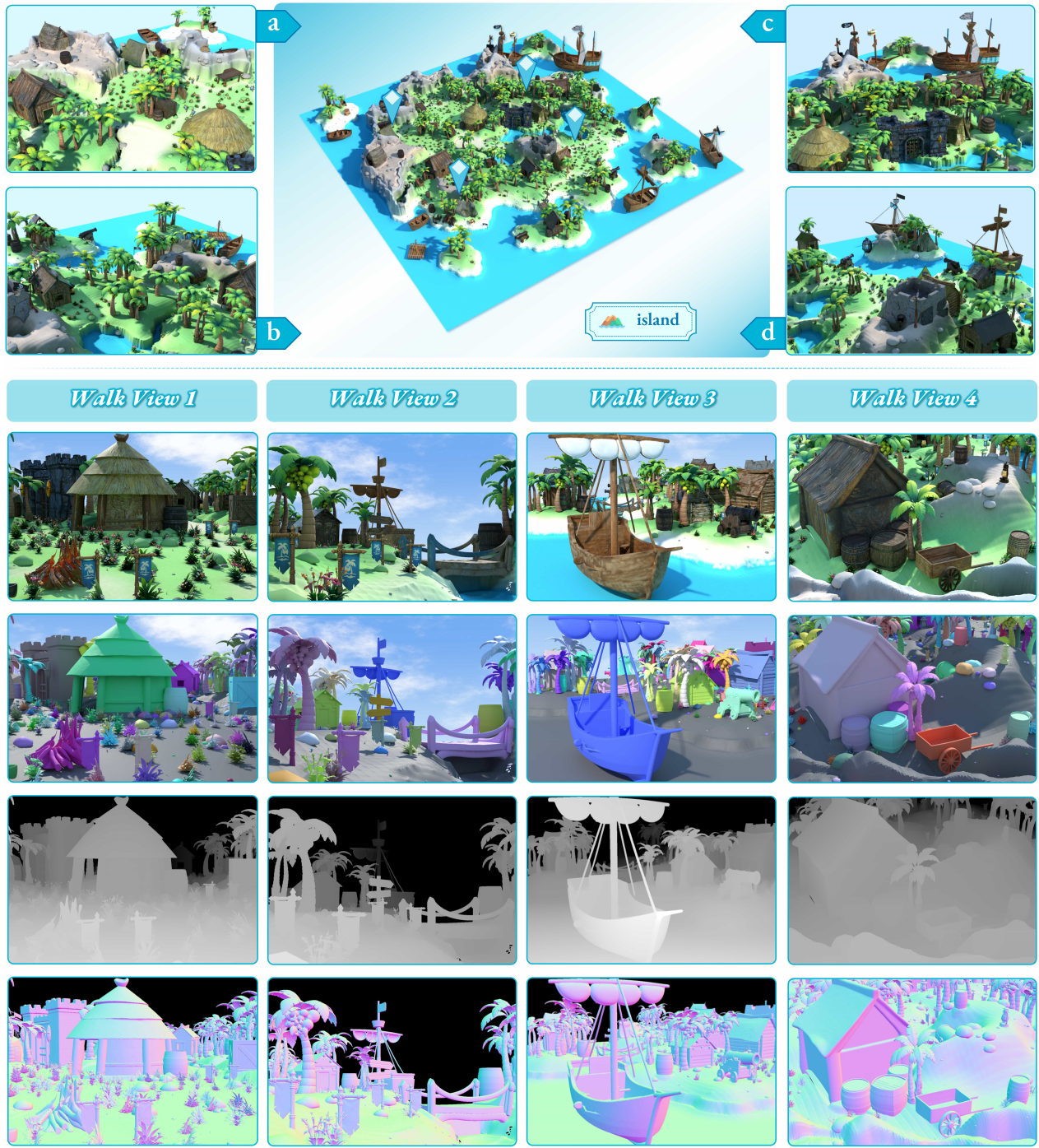}
    \caption{\textbf{Tropical pirate stronghold.}
    An island terrain organizes dense vegetation, settlements,
    docks, and ships into distinct coastal regions. The figure presents the
    global composition, regional close-ups, local walk views, and their
    corresponding instance, depth, and normal renderings.}
    \label{fig:case3_island_small}
\end{figure}

\begin{figure}[H]
    \centering
    \includegraphics[width=1\linewidth]{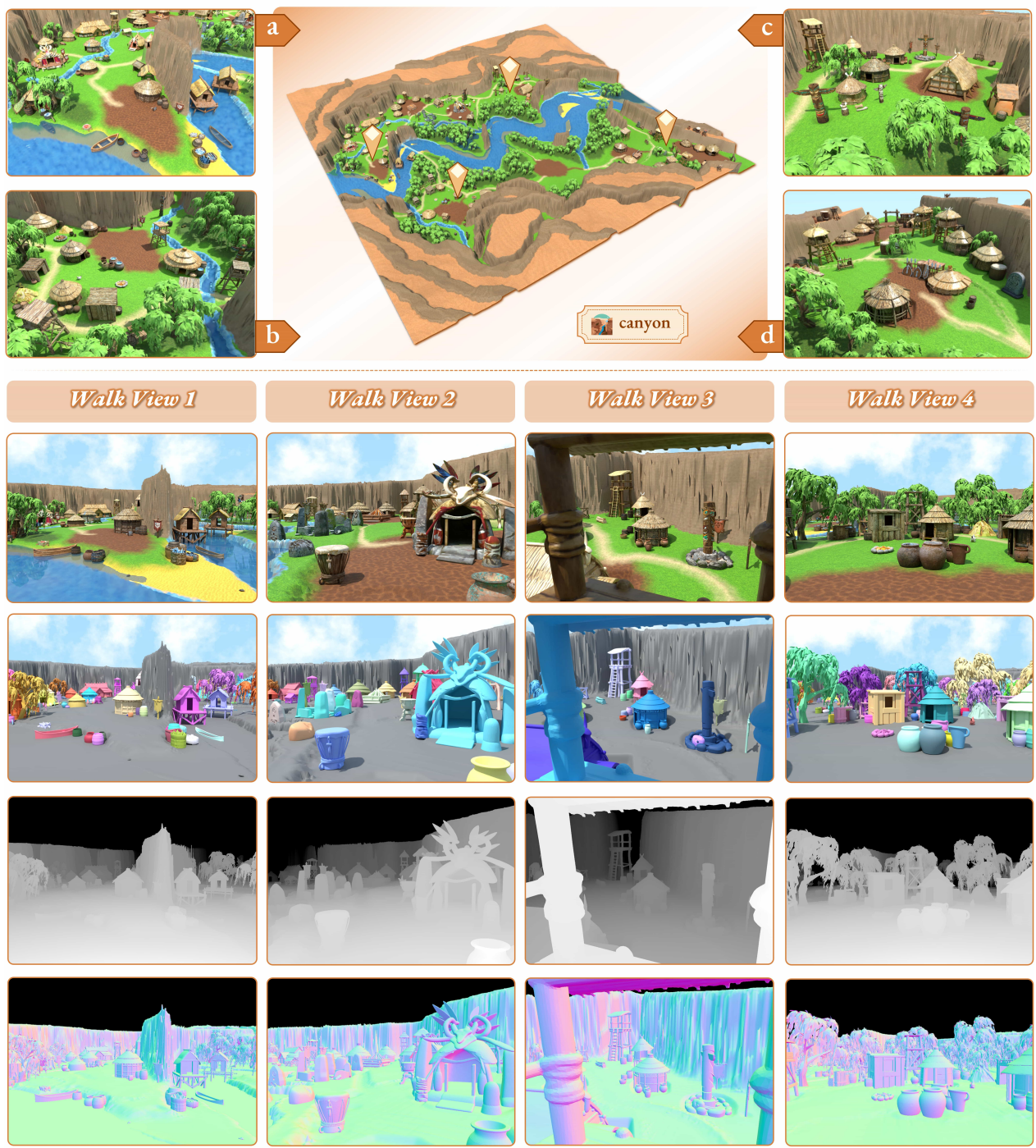}
    \caption{\textbf{Canyon with tribal settlements.}
    A continuous river connects the canyon, valley floor, vegetation, and
    settlement regions across substantial elevation changes. Global and
    regional views are complemented by local walk views and the corresponding
    instance, depth, and normal renderings.}
    \label{fig:case4_canyon_large}
\end{figure}

\begin{figure}[H]
    \centering
    \includegraphics[width=1\linewidth]{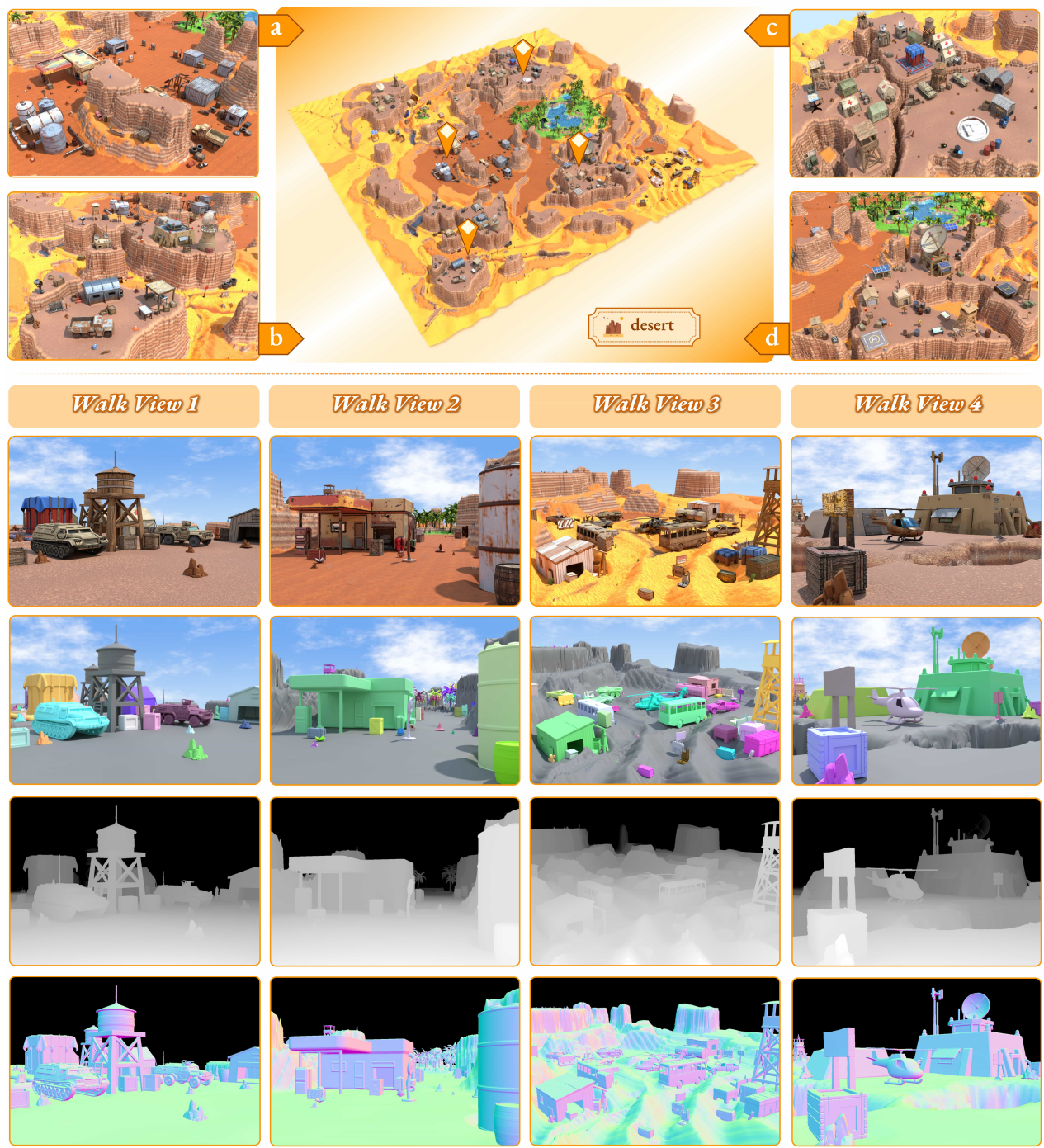}
    \caption{\textbf{Thrilling battlefield in the desert.}
    Layered rocky landforms surround open combat areas and populated compounds
    containing buildings, defensive structures, and vehicles. The figure
    shows the global and regional organization together with local walk views
    and their instance, depth, and normal renderings.}
    \label{fig:case7_desert_large}
\end{figure}

Figures~\ref{fig:case3_island_small}--\ref{fig:case10_snowy_large}
show four representative worlds generated from open-ended prompts: a tropical
pirate island, a river canyon with tribal settlements, a desert battlefield,
and a snow-covered mountain valley with futuristic facilities. The cases span
different terrain structures, scene scales, content densities, and visual
styles. Each figure contains the input prompt, a global view, regional views, and walk views followed by their instance,
depth, and normal rendering images.

The global orbit views highlight the diversity of the generated world
structures. The tropical island contains irregular coastlines and separated
land regions; the canyon is organized around a continuous river and

\begin{figure}[H]
    \centering
    \includegraphics[width=1\linewidth]{figures/case_2/case10.jpg}
    \caption{\textbf{Snow-covered mountain valley with style of Command \& Conquer: Red Alert.}
    The enclosing mountain terrain contains multiple regions populated with
    futuristic facilities, communication structures, and vehicles. Global,
    regional, and local walk views are shown with the corresponding instance,
    depth, and normal renderings.}
    \label{fig:case10_snowy_large}
\end{figure}
\vspace{0.5cm}

substantial elevation variation; the desert combines broad open areas with
layered rocky landforms; and the snow-covered scene forms an enclosed
mountain valley containing multiple functional regions. These examples show
that WorldClaw can express different landform compositions within a coherent
terrain foundation, rather than constructing the world as a collection of
disconnected local scenes.

The local walk views further reveal region-specific content, including settlement buildings, vegetation, docks, defensive structures, communication facilities, and vehicles. Their distributions vary according to the terrain and function of each region, while open areas are retained where required by the scene design. The corresponding instance, depth, and normal renderings expose the
independent scene components, terrain relief, and explicit object surfaces. The final scenes generally maintain plausible object scales, orientations, and object--terrain contacts over both approximately level and sloped surfaces. Together, these visualizations are consistent with the explicit representation of WorldClaw, in which the global terrain and regional objects remain independently manageable textured meshes.

\subsection{Qualitative Comparison}

\begin{figure}[H]
    \centering
    \includegraphics[width=1\linewidth]{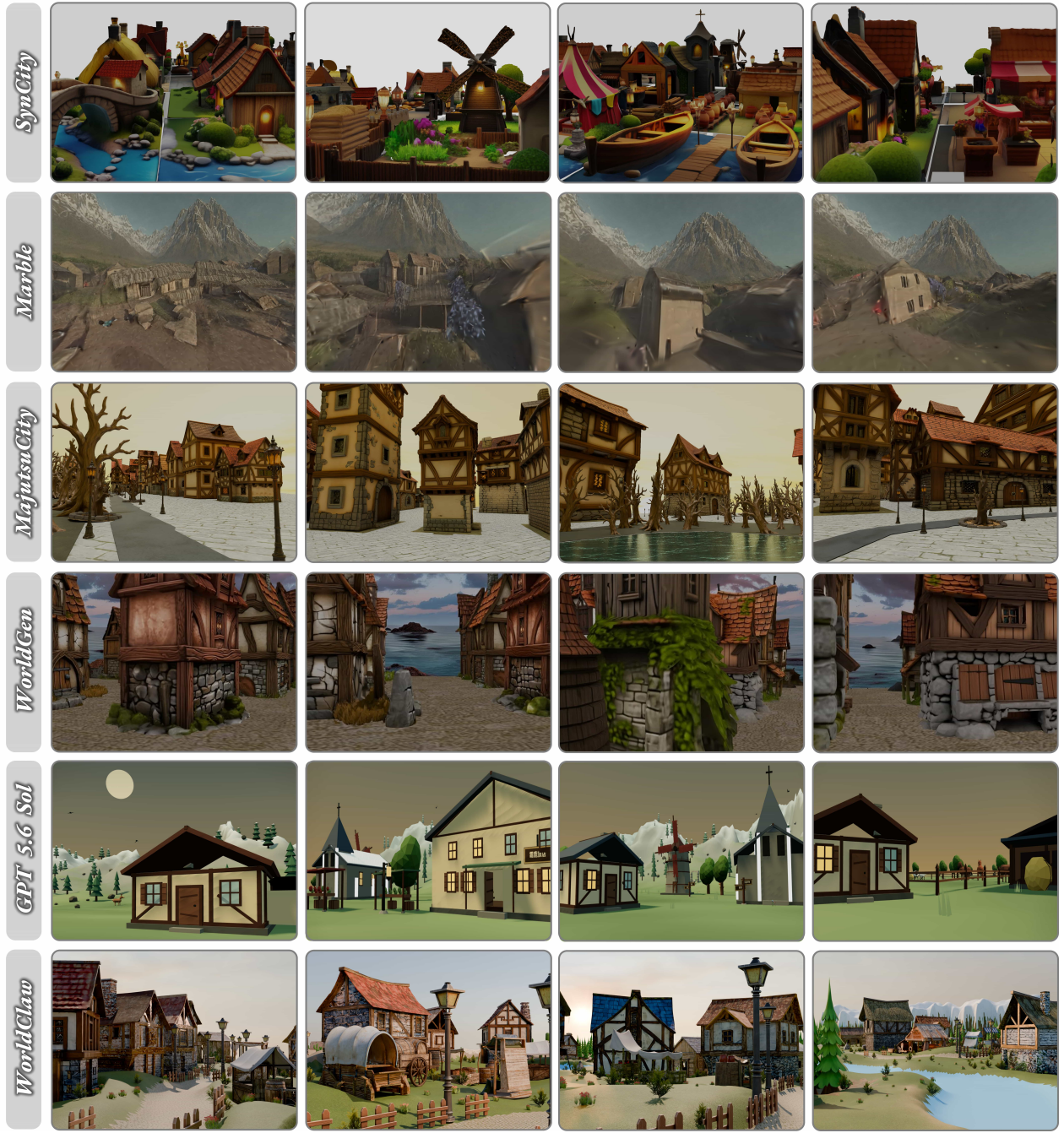}
    \caption{Qualitative comparison with the representative text-driven
    3D scene generation methods. All methods are conditioned on prompts that share the same medieval-village theme and comparable scene requirements, with the wording adapted when necessary to each method's input format. }
    \label{fig:method_comp}
\end{figure}

As shown in Fig.~\ref{fig:method_comp}, we qualitatively compare SynCity~\cite{engstler2025syncity}, Marble~\cite{worldlabs2026marble},
MajutsuCity~\cite{huang2026majutsucity},
WorldGen~\cite{wang2026worldgen}, GPT 5.6 Sol~\cite{openai2026gpt56Sol} and WorldClaw. For each
method, we show four immersive walk views covering different parts of the
generated world.

\paragraph{\textbf{Terrain and Region Organization.}}
We first compare the geometric expressiveness of the generated terrain and the
spatial organization of semantically different regions. SynCity supports
scene-scale synthesis through block-wise 3D latents, although its generated
regions exhibit weaker
long-range organization and less clearly articulated transitions between
terrain types. Marble achieves high visual richness in individual views, but
the displayed terrain lacks explicit region-level organization. MajutsuCity
yields structured building and asset layouts, but its city-oriented design
emphasizes instance arrangement over large-scale landform construction,
leading to comparatively regular ground geometry. WorldGen produces a
coherent and traversable village environment, but its displayed views are
dominated by relatively flat and homogeneous terrain. GPT-5.6 Sol demonstrates
that a coding agent equipped with structured planning and domain-specific
skills can construct a complete terrain-and-object layout. However, its terrain
is realized using relatively simple geometric forms, resulting in less
expressive landforms and coarser transitions between regions. In contrast, WorldClaw
uses a semantic layout map to construct a continuous global terrain, producing
pronounced elevation variation and semantically distinct yet spatially
connected regions.

\paragraph{\textbf{Content Richness and Prompt Alignment.}}
We next assess the richness of the generated scene content and its semantic
alignment with the input prompt. SynCity captures the overall medieval theme,
but some local regions contain relatively coarse or highly repeated
structures, reducing the specificity of the generated content. Marble
produces visually rich individual views, but its content is dominated by
buildings and vegetation, with comparatively few object categories. This
results in lower scene diversity and weaker coverage of the requested
elements. MajutsuCity produces well-arranged buildings and assets, reflecting
its strength in instance-level layout generation; however, its content remains
primarily organized around urban structures and therefore aligns less closely
with the requested village setting. WorldGen generates detailed local
structures and coherent village content, but exhibits relatively limited
content variation across the displayed views. GPT-5.6 Sol captures the main
semantic elements of the prompt, but represents many objects using simplified
or repeated geometry with limited surface detail and stylistic variation.
Although the resulting scene contains the requested content, its assembled,
blockout-like appearance is less visually polished. WorldClaw retains open terrain
where required and selectively introduces settlements, vegetation, animals,
and environmental objects into different regions. This results in richer
content variation and closer alignment between the requested scene elements
and their regional functions.

\paragraph{\textbf{Free-Viewpoint Appearance and Scene Representation.}}
The four ground-level views further reveal differences in geometric stability
and scene representation. SynCity produces a persistent 3D representation,
although its block-wise generation can result in coarse local geometry and
visible inconsistencies between neighboring regions. Marble maintains high
visual fidelity within a limited spatial neighborhood, but its quality
degrades as the camera moves over longer distances. Distant regions exhibit
increasingly incomplete 
or distorted geometry, while close-range views can
expose discrete Gaussian primitives and unstable surfaces. Both SynCity and
Marble represent the generated output primarily at the scene level and do not
expose independently controllable object assets. MajutsuCity retains
independently controllable assets, but its free-viewpoint results are more
strongly shaped by its regular city layouts. WorldGen produces explicit
textured meshes and maintains stable local structures across traversable
views, making it the closest baseline to WorldClaw in terms of downstream
usability. GPT-5.6 Sol also produces explicit 3D geometry and
independently controllable objects. Nevertheless, its simplified object
geometry, basic materials, and limited visual coherence reduce the fidelity
and aesthetic quality of the walk views. WorldClaw combines an explicit global terrain with independently reconstructed and placed object meshes. Consequently, the displayed views preserve both large-scale terrain structure and populated local content while
retaining instance-level editability.

\section{Related Work}

3D scene generation has attracted increasing attention in computer vision and graphics~\cite{wen20253d,tang2025recent,fime2025automatic}. In this section, we review representative studies on static 3D scene generation and categorize existing methods into four paradigms: \textit{procedural generation} (Section \ref{sec:related-procedural}), \textit{visual-prior-based generation} (Section \ref{sec:related-visual}), \textit{native 3D generation} (Section \ref{sec:related-native3d}), and \textit{LLM-driven generation} (Section \ref{sec:related-llm_driven}).

\subsection{Procedural 3D Scene Generation}
\label{sec:related-procedural}
Procedural generation synthesizes scene content through predefined algorithms, rules, and executable programs, and has long served as a fundamental paradigm for virtual environment creation. According to the evolution of generation paradigms, existing procedural approaches can be broadly categorized into classical procedural generation and LLM-guided procedural generation.

\textbf{Classical procedural generation.}
Classical procedural methods construct virtual environments by explicitly encoding generation rules, stochastic models, or grammar-based algorithms. They have been extensively applied to the synthesis of terrains~\cite{musgrave1989synthesis,genevaux2013terrain}, vegetation~\cite{weber1995creation,deussen1998realistic}, and buildings and cities~\cite{parish2001procedural,wonka2003instant,muller2006procedural}, and have become a standard content-generation technique in games, films, simulation, and virtual-world creation~\cite{hendrikx2013procedural,smelik2014survey}. Owing to their controllability, scalability, and repeatability, procedural methods are also well suited for constructing synthetic datasets with automatically generated geometry, materials, and annotations for computer vision. Representative examples include the Infinigen series: Infinigen~\cite{raistrick2023infinite}, which procedurally synthesizes natural environments containing terrains, vegetation, animals, and materials; Infinigen Indoors~\cite{raistrick2024infinigen}, which extends the framework to complete indoor environments; and Infinigen-Articulated~\cite{joshi2026procedural}, which further introduces articulated assets with physical properties and simulator-ready annotations.

\textbf{LLM-guided procedural generation.}
Recent advances in large language models (LLMs) and multimodal foundation models have enabled procedural generation to be driven by natural language and visual conditions. Rather than manually specifying procedural parameters, these methods employ LLMs to interpret user intent, organize generation workflows, and coordinate procedural modules. CityX~\cite{zhang2024cityx} and SceneX~\cite{zhou2025scenex} utilize LLM planners and multi-agent systems to orchestrate procedural pipelines for large-scale urban scenes and natural environments. LatticeWorld~\cite{duan2025latticeworld} translates multimodal instructions into symbolic layouts and environmental configurations that are executed in Unreal Engine, while LandCraft~\cite{liu2026landcraft} combines semantic layouts, height maps, and procedural terrain modules to synthesize structured landscapes. Compared with classical procedural generation, these methods substantially improve semantic controllability and automation. Nevertheless, scene synthesis still fundamentally depends on predefined procedural operators and handcrafted generation modules, limiting the flexibility and generality of procedural world construction.

\subsection{Visual-Prior-Based 3D Scene Generation}
\label{sec:related-visual}
Visual-prior-based methods leverage pretrained image or video generative models to synthesize visual observations and lift them into 3D through depth estimation, geometric reprojection, multi-view fusion, or explicit representation optimization. Based on their dominant scene-construction strategies, existing methods can be broadly organized into progressive scene expansion, global scene initialization, structured scene generation, and video-guided world generation.

\textbf{Progressive scene expansion.}
One line of research progressively synthesizes and integrates visual observations along camera trajectories to extend scene coverage and construct an explorable 3D environment. Infinite Nature~\cite{liu2021infinite}, InfiniteNature-Zero~\cite{li2022infinitenature}, Persistent Nature~\cite{chai2023persistent}, and DiffDreamer~\cite{cai2023diffdreamer} primarily focus on extending natural scenes over long-range camera motion. SceneScape~\cite{fridman2023scenescape} combines text-conditioned view synthesis with monocular depth estimation to generate geometrically consistent long-term views, while WonderJourney~\cite{yu2024wonderjourney} constructs sequences of diverse yet semantically connected point-cloud scenes. ScenePainter~\cite{xia2025scenepainter} further improves long-range consistency by mitigating semantic drift during successive scene expansion.

Other methods following this incremental construction paradigm place greater emphasis on integrating generated observations into a persistent scene-level representation. Text2Room~\cite{hollein2023text2room} progressively synthesizes RGB-D observations from selected viewpoints and fuses the resulting geometry into a seamless textured mesh of an indoor environment. LucidDreamer~\cite{chung2023luciddreamer} alternates between image inpainting, depth-based point-cloud expansion, and geometric alignment before optimizing the accumulated observations as a unified 3D Gaussian scene. RealmDreamer~\cite{shriram2024realmdreamer} similarly constructs a forward-facing 3D Gaussian representation using diffusion-based inpainting and monocular depth supervision. WonderTurbo~\cite{ni2025wonderturbo} accelerates incremental geometry and appearance construction through efficient depth completion, Gaussian updates, and lightweight diffusion inpainting. Beyond individual environments, WonderWorld~\cite{yu2025wonderworld} enables low-latency, user-controlled expansion of connected 3D scenes, whereas Infinicity~\cite{lin2023infinicity}, CityDreamer~\cite{xie2024citydreamer}, and GaussianCity~\cite{xie2025generative} extend generative scene construction toward large-scale urban environments.

\textbf{Global scene initialization.}
Rather than constructing a scene solely through sequential extrapolation, another line of work first establishes a globally coherent visual proxy, typically in the form of an omnidirectional panorama, and subsequently lifts it into 3D. PERF~\cite{wang2024perf} represents an early attempt to reconstruct a panoramic neural radiance field from a single panorama by combining panoramic depth estimation with progressive RGB-D inpainting. More recent methods combine generative panoramic priors with efficient 3D Gaussian Splatting. FastScene~\cite{ma2024fastscene} generates a text-conditioned indoor panorama, synthesizes additional observations through progressive inpainting, and reconstructs the environment using panoramic Gaussian splatting. DreamScene360~\cite{zhou2024dreamscene360} generates a globally coherent panorama and lifts it into an unconstrained 3D Gaussian scene. HoloDreamer~\cite{zhou2024holodreamer} employs a high-resolution equirectangular panorama as a holistic scene initialization and adopts a two-stage Gaussian reconstruction process to produce view-consistent, enclosed 3D worlds. SceneDreamer360~\cite{li2024scenedreamer360} further improves text-conditioned panoramic generation and introduces point-cloud-based initialization for spatially consistent panoramic Gaussian reconstruction.

Beyond direct panoramic lifting, LayerPano3D~\cite{yang2025layerpano3d} and Scene4U~\cite{huang2025scene4u} decompose panoramic observations into multiple depth or semantic layers to recover occluded content and obtain more structured scene representations. HunyuanWorld 1.0~\cite{team2025hunyuanworld} further integrates panoramic generation, geometric reconstruction, and semantic scene decomposition into a unified framework for producing explorable and structured 3D environments. Global initialization provides a more coherent scene-level prior and reduces the long-range drift associated with sequential extrapolation.

\textbf{Structured scene generation.}
More recent methods seek to construct scenes as collections of semantically meaningful instances or independently manageable 3D components, rather than representing the entire environment as a single mesh or 3D Gaussian. General image-to-3D models such as SAM3D~\cite{chen2026sam3d} provide open-set object reconstruction priors that can be applied to individual scene elements. MIDI~\cite{huang2025midi} extends pretrained object-generation priors to jointly generate multiple 3D instances while modeling their interactions and spatial relationships. SceneTransporter~\cite{wang2026scenetransporter} similarly recovers multiple scene entities and their structural arrangements from visual observations.

A modular line of research explicitly decomposes scene construction into object-level stages. SceneMaker~\cite{shi2026scenemaker} separates object de-occlusion, 3D generation, and pose estimation to improve open-set scene construction under severe occlusion. CAST~\cite{yao2025cast} and related approaches recover individual objects together with their spatial configurations, while Sketch2Scene~\cite{xu2024sketch2scene} extracts semantic components and layout information from a generated isometric image and instantiates the resulting scene through a procedural game-engine pipeline. Other methods incorporate hierarchical parsing, background reconstruction, and structural or physical constraints to obtain independently editable or replaceable scene components~\cite{dong2025hiscene,sautter20263d}. WorldGen~\cite{wang2026worldgen} combines language-based layout reasoning, procedural blockouts, navigation constraints, reference-image generation, holistic reconstruction, scene decomposition, and object-level enhancement to produce traversable and editable 3D worlds compatible with standard game engines. Compared with monolithic scene representations, these methods provide stronger support for instance-level manipulation and downstream interaction.

\textbf{Video-guided world generation.}
Video generative models provide temporally coherent visual observations under continuous camera motion, offering denser multi-view evidence for static 3D scene construction. Scene Splatter~\cite{zhang2025scene} uses video diffusion to synthesize consistent novel views and iteratively refine a global 3D Gaussian representation, whereas Wonderland~\cite{liang2025wonderland} directly predicts a 3D scene representation from camera-conditioned video latents. To enlarge the range of exploration, WorldExplorer~\cite{schneider2025worldexplorer} coordinates multiple autoregressive camera trajectories using persistent scene memory, while FlexWorld~\cite{chen2026flexworld} progressively integrates generated observations through geometry-aware fusion. More recently, HY-World 2.0~\cite{hy2026hy} unifies video-based world expansion with feed-forward 3D reconstruction in a multi-modal world model, using video diffusion to generate extended scene observations and WorldMirror 2.0~\cite{liu2025worldmirror} to recover geometrically consistent 3D world representations from multi-view images or videos.

\subsection{Native 3D Scene Generation}
\label{sec:related-native3d}
Native 3D methods directly model scene geometry or its latent 3D representations without relying on lifting independently generated visual observations, thereby avoiding the geometric ambiguities and structural inconsistencies introduced by image- or video-based reconstruction. According to their scene generation strategies, existing methods can be broadly categorized into holistic scene generation, progressive scene expansion, and synchronized scene generation.

\textbf{Holistic scene generation.}
Early native 3D methods directly synthesize complete scenes within a predefined spatial extent. DiffInDScene~\cite{ju2024diffindscene} generates global occupancy using cascaded sparse diffusion and reconstructs indoor details through overlapping TSDF blocks. XCube~\cite{ren2024xcube} adopts hierarchical sparse-voxel diffusion to improve scene resolution and semantic attribute capacity. Frankenstein~\cite{yan2024frankenstein} jointly decodes multiple semantic signed distance fields from a shared tri-plane representation, preserving separable indoor components, while L3DG~\cite{roessle2024l3dg} introduces diffusion in a compressed 3D Gaussian latent space for renderable room generation. For outdoor environments, CADD~\cite{hu2025large} and CymbaDiff~\cite{liang2026cymbadiff} synthesize semantic scene volumes through multi-scale discrete diffusion, sparse absorbing refinement, and structured spatial sequence modeling. More recently, NuiWorld~\cite{lee2026nuiworld} represents varying numbers of scene chunks as variable-length latent sequences, enabling more flexible scene extents while still generating all chunks jointly within a single global generative process. Although these methods produce globally consistent scene representations, their generation capacity remains coupled to a predefined spatial support.

\textbf{Progressive scene expansion.}
Progressive methods relax the fixed-scene assumption by formulating scene generation as conditional 3D continuation. Starting from a seed region, they repeatedly synthesize neighboring scene content conditioned on previously generated geometry. BlockFusion~\cite{wu2024blockfusion} constrains overlapping latent tri-planes with known scene content throughout reverse diffusion, enabling iterative extrapolation of new mesh blocks. SceneFactor~\cite{bokhovkin2025scenefactor} decomposes scene generation into semantic latent outpainting followed by geometry refinement through unsigned distance fields, while SemCity~\cite{lee2024semcity} progressively extends outdoor semantic voxel fields via tri-plane manipulation. NuiScene~\cite{lee2025nuiscene} instead explicitly learns conditional distributions between neighboring vector-set scene chunks, avoiding repeated diffusion resampling during inference. WorldGrow~\cite{li2026worldgrow} further introduces object-level structured latents and performs coarse-to-fine 3D block inpainting to jointly generate geometry and appearance. GaussianGPT~\cite{von2026gaussiangpt} serializes quantized 3D Gaussians into autoregressive tokens, supporting unified scene generation, completion, and continuation. Although progressive generation enables scalable scene expansion, its locally conditioned synthesis remains generation-order dependent and may accumulate layout drift and long-range semantic inconsistencies.

\textbf{Synchronized scene generation.}
Rather than generating scene blocks sequentially, synchronized methods jointly optimize multiple overlapping regions to improve global consistency. LT3SD~\cite{meng2025lt3sd} establishes coarse global structures through latent-tree outpainting before jointly denoising overlapping latent patches at finer resolutions. TRELLISWorld~\cite{chen2025trellisworld} applies object-level 3D diffusion priors across all overlapping latent tiles and aggregates their predictions throughout the denoising process. Extend3D~\cite{yoon2026extend3d} enlarges the latent space of pretrained object models and generates larger scenes by coupling the denoising trajectories of neighboring subregions. WorldFlow3D~\cite{joshi2026worldflow3d} instead formulates scene generation as hierarchical flow matching from coarse geometry to fine geometry and appearance, synchronizing overlapping volumetric regions through merged velocity fields. Compared with progressive expansion, synchronized generation substantially improves global coherence and boundary consistency between neighboring regions, but typically requires the target scene extent to be specified in advance, with computation and memory consumption increasing rapidly as scene scale grows.

\subsection{LLM-Driven 3D Scene Generation}
\label{sec:related-llm_driven}
LLM-driven methods leverage large language models to interpret user intent, reason about spatial relationships, generate executable scene representations, and orchestrate downstream generation tools. According to the evolution of system capabilities, existing approaches can be broadly categorized into language-guided scene planning, hierarchical and interactive scene construction, and agentic scene generation.

\textbf{Language-guided scene planning.}
Early work primarily employs LLMs to translate natural language into scene layouts, object relationships, or executable scene descriptions. Holodeck~\cite{yang2024holodeck} and I-Design~\cite{ccelen2024design} infer object sets and spatial relationships from user instructions, followed by asset retrieval and geometric layout optimization. 3D-GPT~\cite{sun20253d} converts textual descriptions into procedural functions and parameterized scene representations, while SceneCraft~\cite{hu2024scenecraft} further compiles relational scene graphs into executable Blender Python programs. These methods enable natural language to directly participate in high-level scene planning and procedural execution, but typically rely on one-shot planning with limited iterative refinement.

\textbf{Hierarchical and interactive scene construction.}
Subsequent research extends LLM-driven generation toward larger-scale planning, richer scene hierarchies, and visually grounded interaction. HSM~\cite{pun2026hsm} and OptiScene~\cite{yang2026optiscene} improve hierarchical indoor scene organization and preference-aware layout generation. CityCraft~\cite{deng2024citycraft}, MajutsuCity~\cite{huang2026majutsucity}, Yo’City~\cite{lu2026yo}, and MANSION~\cite{che2026mansion} further expand LLM planning to city-scale environments, architectural styles, functional zoning, and multi-floor buildings. Meanwhile, SceneThesis~\cite{ling2025scenethesis}, spatially contextualized VLM agents~\cite{liu2025agentic}, and HOLODECK 2.0~\cite{bian2025holodeck} incorporate image guidance, structured spatial memory, and interactive editing, gradually transforming LLM-driven scene generation from one-shot layout prediction toward visually grounded and iterative scene construction. Although these methods significantly improve planning quality and interaction capability, scene generation is still largely constrained by predefined generation pipelines and limited feedback mechanisms.

\textbf{Agentic scene generation.}
More recent work formulates 3D scene generation as an observable, executable, and self-correcting agentic process. SceneWeaver~\cite{yang2026sceneweaver}, SAGE~\cite{xia2026sage}, and SceneSmith~\cite{pfaff2026scenesmith} iteratively modify scenes through tool invocation, visual feedback, and physical verification. VIGA~\cite{yin2026vision} adopts a code–execute–render–inspect loop for long-horizon optimization of graphics programs, while SceneCode~\cite{wang2026scenecode} compiles user requests into executable scene-generation programs and supports local regeneration through code repair and render-based verification. By explicitly integrating reasoning, execution, observation, and feedback, these agent-based systems substantially improve the autonomy and robustness of scene generation.

\section{Limitations}
\label{sec:limitation_futurework}

While WorldClaw demonstrates the potential of agentic pipelines for open-world 3D scene generation, several limitations remain.

\textbf{High dependency on underlying models.}
WorldClaw decomposes world construction into multiple stages coordinated by heterogeneous models: large language models perform planning and procedural design, image generation models produce semantic layouts and regional object compositions, and 3D generation models reconstruct individual assets. Although this decoupling improves stage-wise controllability, it places strong demands on the generalization capabilities of the underlying models. In our experiments, current open-source language models often struggled to generate procedural terrain and materials that were both executable and consistent with user requirements. Likewise, open-source image generation models frequently failed to produce usable semantic layout maps or to preserve object appearance and pose during object-image generation and extraction. The visual quality of the final scene is also directly bounded by the 3D generation backbone, as low-fidelity geometry and textures noticeably reduce immersion. Consequently, fully validating this decoupled pipeline at the current stage still requires capable models such as Claude Opus 4.8, GPT-Image-2, and Hunyuan3D.

\textbf{Stability risks in code generation.}
Several stages of WorldClaw rely on LLM-generated programs for terrain construction, procedural material generation, asset placement, and local refinement. However, consistently translating high-level natural-language requirements into concrete programs remains difficult. Errors in scale estimation, numerical parameters, or node connectivity directly manifest in the resulting 3D scene as inconsistent landforms, inaccurate material effects, or object layouts that deviate from the user intent, often necessitating multiple render--inspect--refine iterations. This issue is particularly evident for professional 3D creation software such as Blender, whose APIs and complex node-based workflows remain challenging for current language models. Consequently, material effects that artists can construct with sophisticated node graphs are often reduced to relatively simple approximations in generated programs.

\textbf{Efficiency overhead in scene generation.}
To construct richly populated 3D scenes with instance-level editability, WorldClaw generates and reconstructs individual objects separately and performs multiple rounds of agentic refinement over the terrain, assets, and their contact relationships. This long-horizon pipeline incurs substantial inference latency and computational cost, both of which increase with the number of objects and refinement iterations. Although this overhead provides additional editability for complex scenes that require fine-grained control, the pipeline can be unnecessarily lengthy and inefficient for simpler scenes that holistic generation methods can synthesize in fewer steps. This efficiency limitation becomes more pronounced when constructing larger or denser 3D worlds.

\section{Conclusion}
\label{sec:conclusion}

In this work, we presented \textbf{WorldClaw}, a coarse-to-fine agentic framework for generating explicit, explorable, and editable 3D worlds from open-ended text prompts. The central design of WorldClaw is to decouple global world organization from local instance-level content generation. The system first translates user intent into a structured scene specification and constructs a global terrain with region-aware semantics and multi-scale landform structures. It then selectively generates and places independent 3D objects in regions that require fine-grained content, enriching local scene composition while preserving the established global spatial organization. Render-guided agents further refine terrain appearance, object quality, spatial arrangements, and object--terrain contacts. The resulting worlds consist of explicit terrain and independently manageable textured meshes, supporting free-viewpoint exploration, object-level editing and reuse, and integration with conventional rendering, animation-authoring, and game-engine workflows. Our experiments demonstrate that WorldClaw can generate 3D worlds with diverse landforms, clear regional organization, and rich object composition from open-ended prompts, while achieving an effective balance among scene scale, visual quality, and editability.

The current framework still relies primarily on generative 3D models to reconstruct regional objects. Although such models provide substantial geometric and appearance diversity, they do not consistently recover explicit part hierarchies, parametric structures, articulation definitions, or interaction logic. As foundation models continue to improve in code generation, spatial reasoning, and tool use, code-native 3D modeling offers a complementary direction for addressing these limitations~\cite{lu2025ll3m, zhou2026articraft, gao20263dcodebench, yang2026p3d}. In WorldClaw, we have already explored this direction for terrain appearance generation, where the agent constructs terrain materials through executable procedures such as Blender material-node graphs and shader scripts. These code-driven materials achieve encouraging visual results while remaining explicitly parameterized and readily editable, suggesting that executable programs can effectively represent not only object geometry, but also complex scene appearance. Future work may extend this strategy to replace selected object-generation stages with executable modeling programs, or combine programmatic structure and material construction with generative geometry and appearance priors. Compared with approaches that produce fixed meshes or textures alone, executable representations can explicitly encode object composition, material logic, adjustable parameters, and motion constraints, thereby improving the controllability, reusability, and animatability of generated assets. As the coding capabilities of foundation models continue to advance, we believe that increasingly complete immersive 3D worlds may eventually be constructed primarily, or even entirely, through executable code.

Another important direction is to integrate WorldClaw with more mature production-oriented scene-construction and procedural-content-generation tools. Our current implementation primarily relies on Blender, which provides convenient scriptable access to geometry, materials, and rendering, but large-scale game-world construction additionally requires runtime systems for procedural generation, navigation, physics, and interaction. Game engines such as Unreal Engine provide a more comprehensive ecosystem of procedural-content-generation tools and real-time scene capabilities. Recent efforts that construct generative worlds using game-engine procedural tools further indicate the potential of this direction~\cite{liu2026simworlds, kang2026simworld}. Integrating the planning and code-generation capabilities of WorldClaw with procedural generation, level editing, shader authoring, physical simulation, and runtime interaction systems in such engines could improve the efficiency, scalability, and practical applicability of large-scale 3D world generation.

More broadly, we envision WorldClaw as a step toward extending generative visual intelligence beyond static output synthesis to the construction of executable, editable, and production-ready 3D content. By combining explicit 3D representations, programmatic geometry and material modeling, and agentic tool use, future generative systems may produce not only the visual appearance of a scene, but also complete 3D worlds whose geometry, materials, internal structures, animation capabilities, and interactive behaviors are jointly defined through executable representations. Such systems could further narrow the gap between open-ended generative models and practical 3D content-production workflows.

Looking further ahead, we hope that world creation will no longer be constrained by the need to manually prepare scene assets, material libraries, or shader programs. Creators would no longer need to devote substantial effort to searching for suitable models, constructing complex material graphs, or implementing specific visual effects. Instead, these technical processes could be handled automatically by agents equipped with code-generation, visual-understanding, and tool-use capabilities. At that point, the central question of 3D content creation would no longer be how to construct every underlying component, but what kind of world the creator wishes to express. We would only need to focus on imagining and creating a world that is uniquely our own---and nothing more.

\section{Additional Results}

\begin{figure}[ht]
    \centering
    \includegraphics[width=1\linewidth]{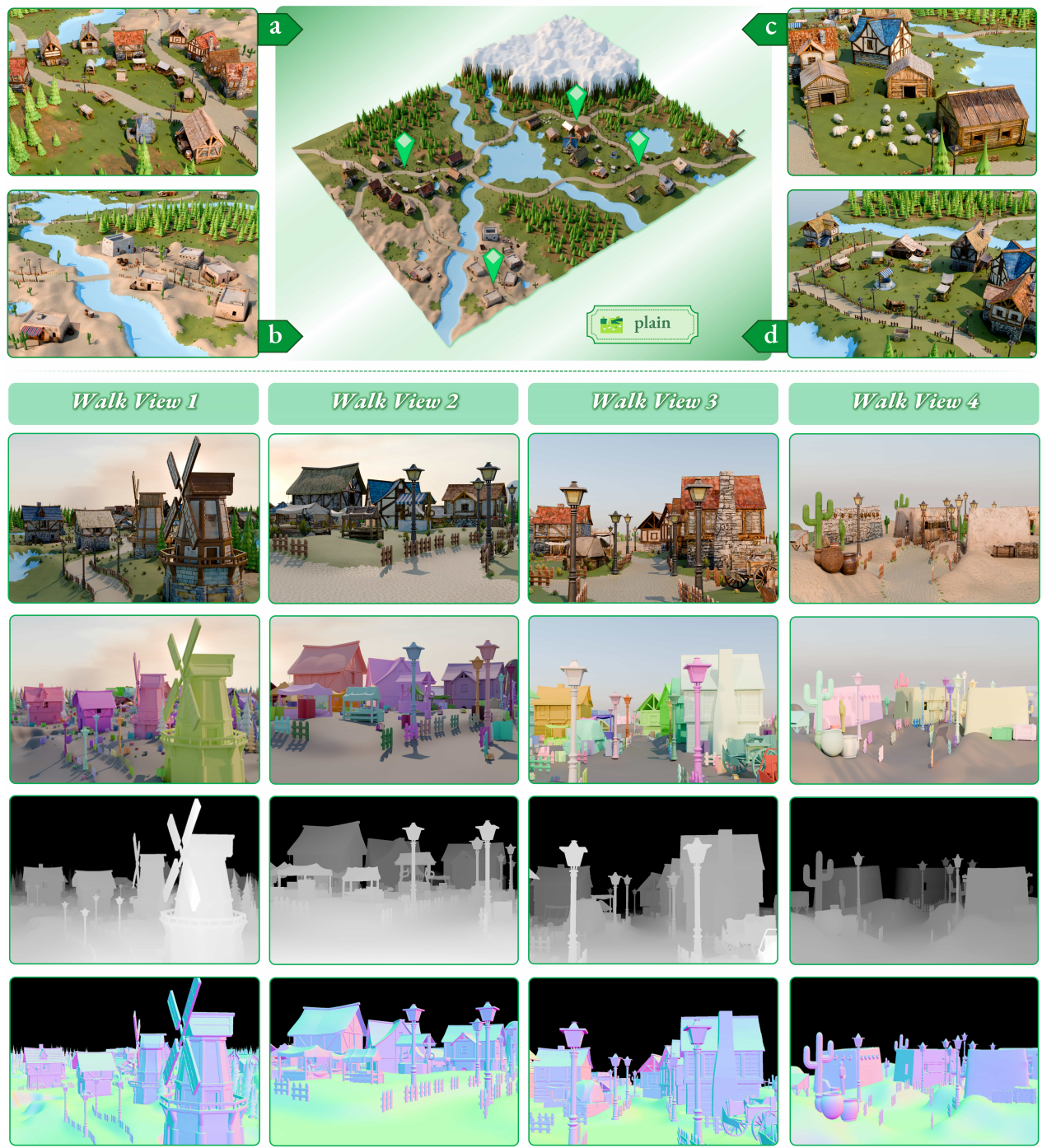}
    \caption{\textbf{Medieval-style village across diverse terrain.}
    The scene combines snow-capped mountains, forested plains, waterways, and
    desert areas, with village buildings, windmills, vegetation, and animals
    distributed across the different regions. The figure shows the global
    layout, regional close-ups, local walk views, and their corresponding
    instance, depth, and normal renderings.}
    \label{fig:case0_plain_small}
\end{figure}

\begin{figure}[ht]
    \centering
    \includegraphics[width=1\linewidth]{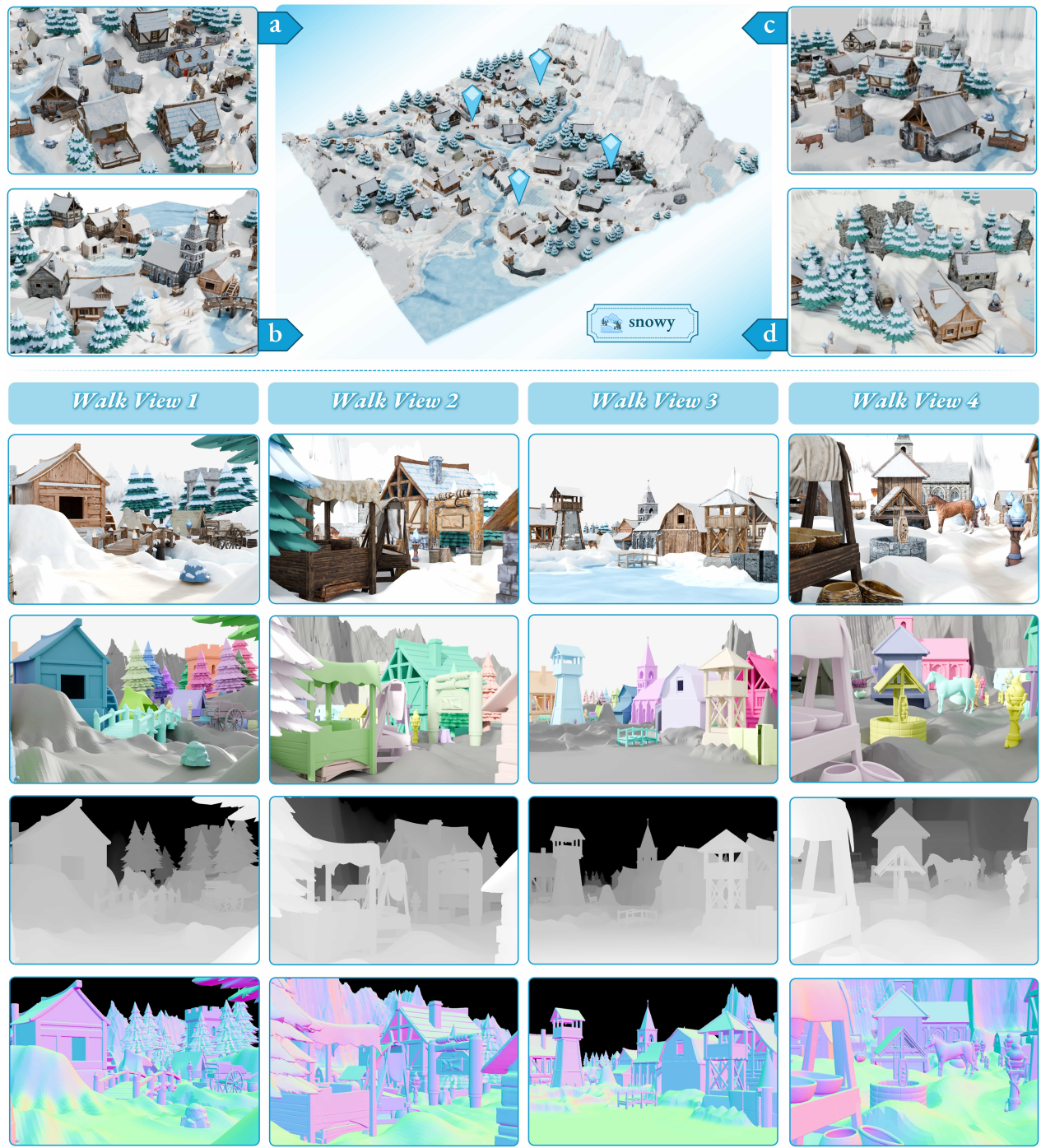}
    \caption{\textbf{Snow-covered riverside village.}
    The village extends along both sides of a frozen river within a mountainous
    landscape. The figure shows the global layout, regional close-ups, and
    local walk views together with instance, depth, and normal renderings.}
    \label{fig:case1_snowy_small}
\end{figure}

\begin{figure}[ht]
    \centering
    \includegraphics[width=1\linewidth]{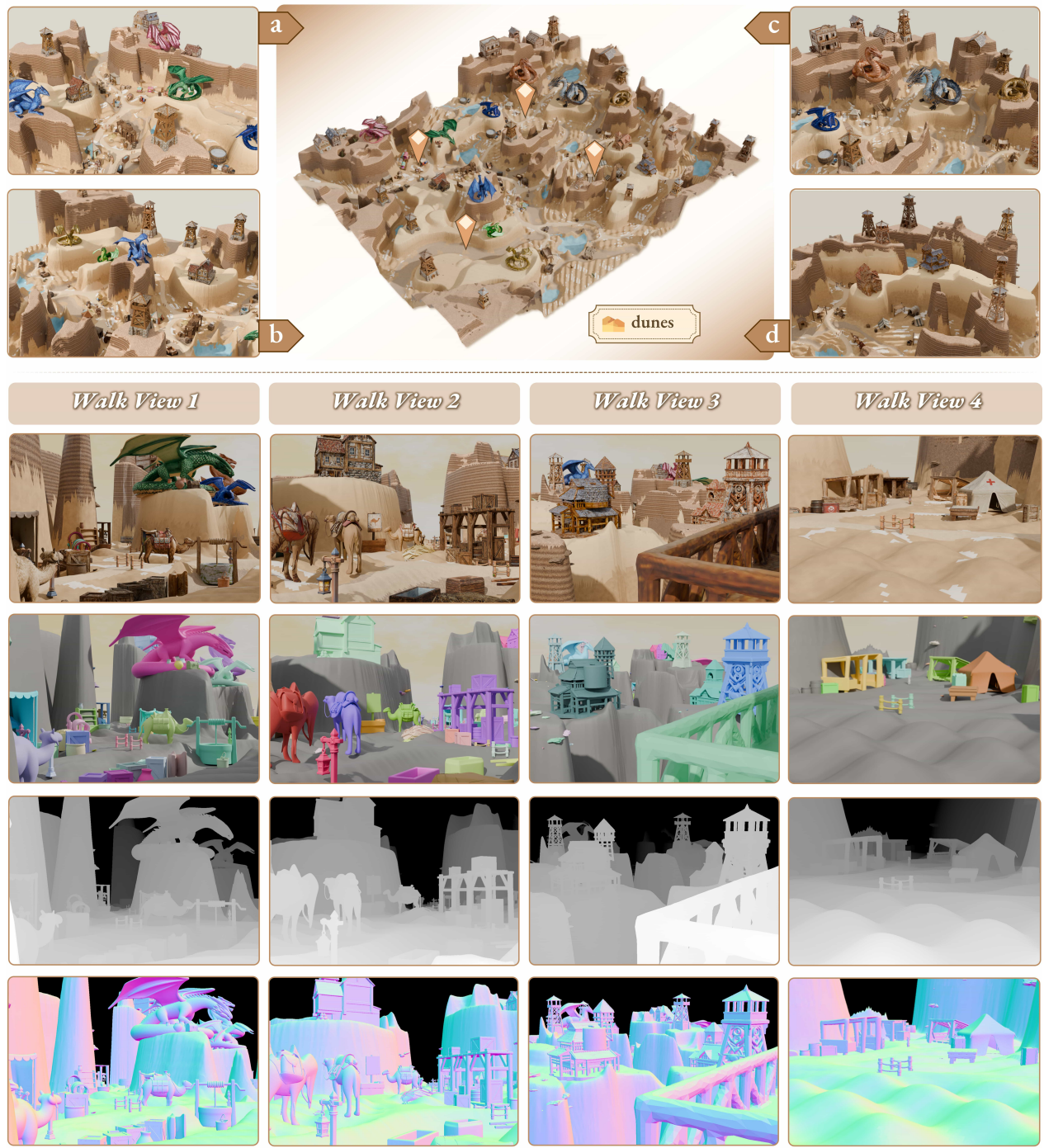}
    \caption{\textbf{Desert adventure camp surrounded by dragons.}
    The scene combines layered desert terrain, settlements, watchtowers, and
    dragons distributed around the camp. Global, regional, and local walk
    views are shown with the corresponding instance, depth, and normal
    renderings.}
    \label{fig:case2_dunes_small}
\end{figure}

\begin{figure}[ht]
    \centering
    \includegraphics[width=1\linewidth]{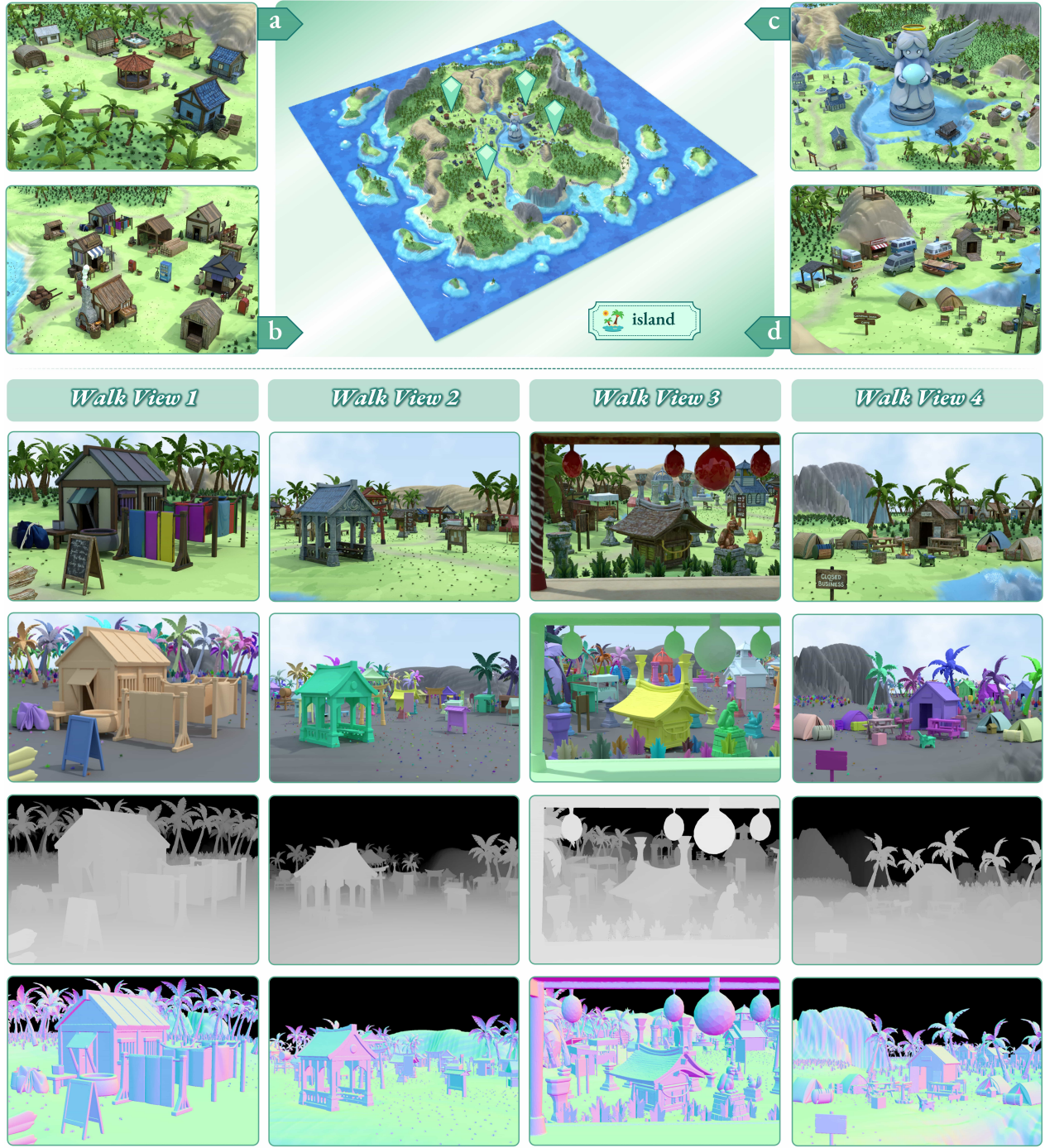}
    \caption{\textbf{Island with Japanese-style towns.}
    Multiple settlements are distributed across an island containing
    coastlines, vegetation, hills, and water. The figure presents the global
    and regional organization as well as local walk views and their instance,
    depth, and normal renderings.}
    \label{fig:case5_island_large}
\end{figure}

\begin{figure}[ht]
    \centering
    \includegraphics[width=1\linewidth]{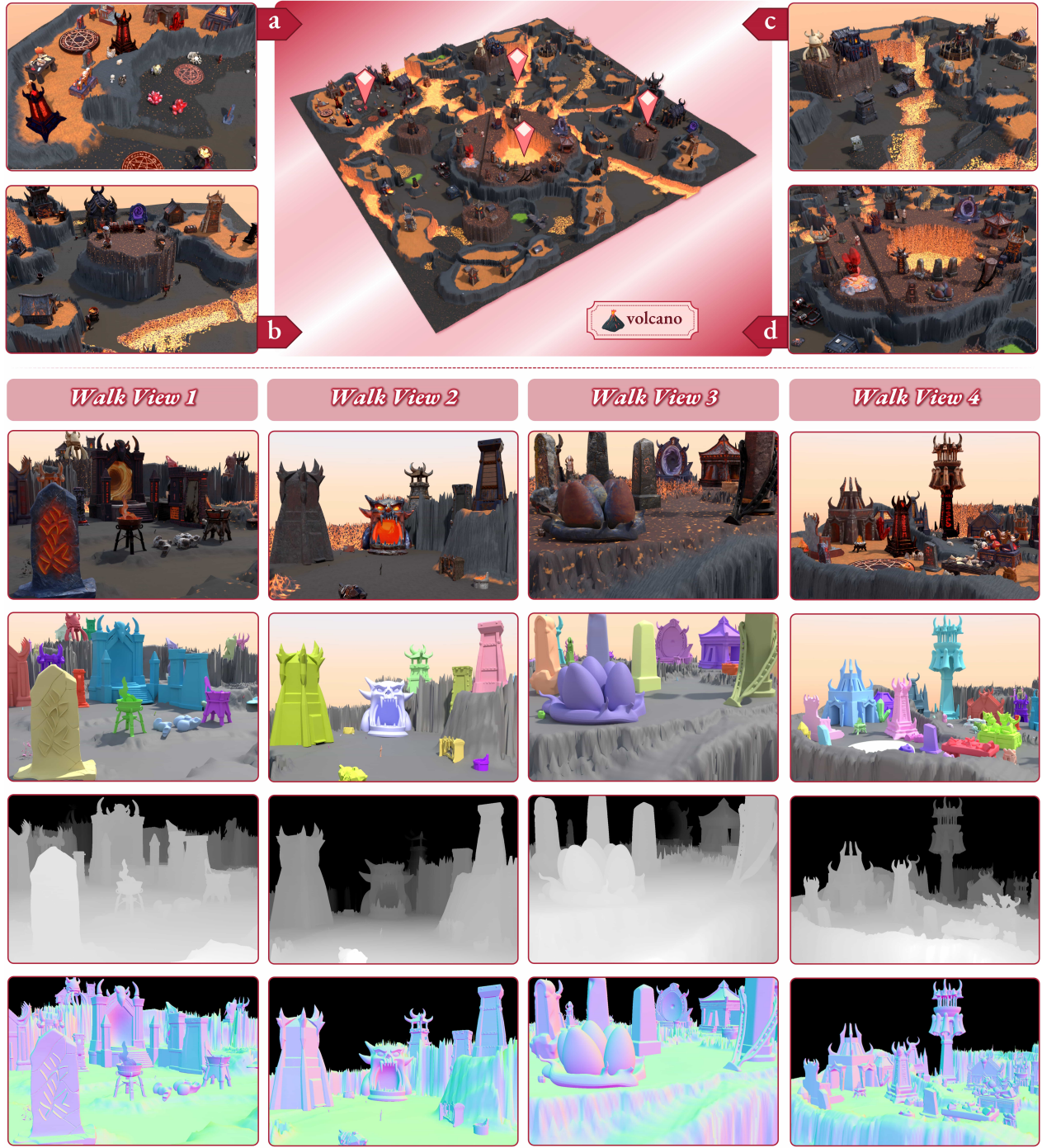}
    \caption{\textbf{Volcanic demon lair.}
    Lava flows, volcanic landforms, and demonic structures form a unified
    environment around the central lair. Global and regional views are
    complemented by local walk views and the corresponding instance, depth,
    and normal renderings.}
    \label{fig:case6_volcano_large}
\end{figure}

\begin{figure}[ht]
    \centering
    \includegraphics[width=1\linewidth]{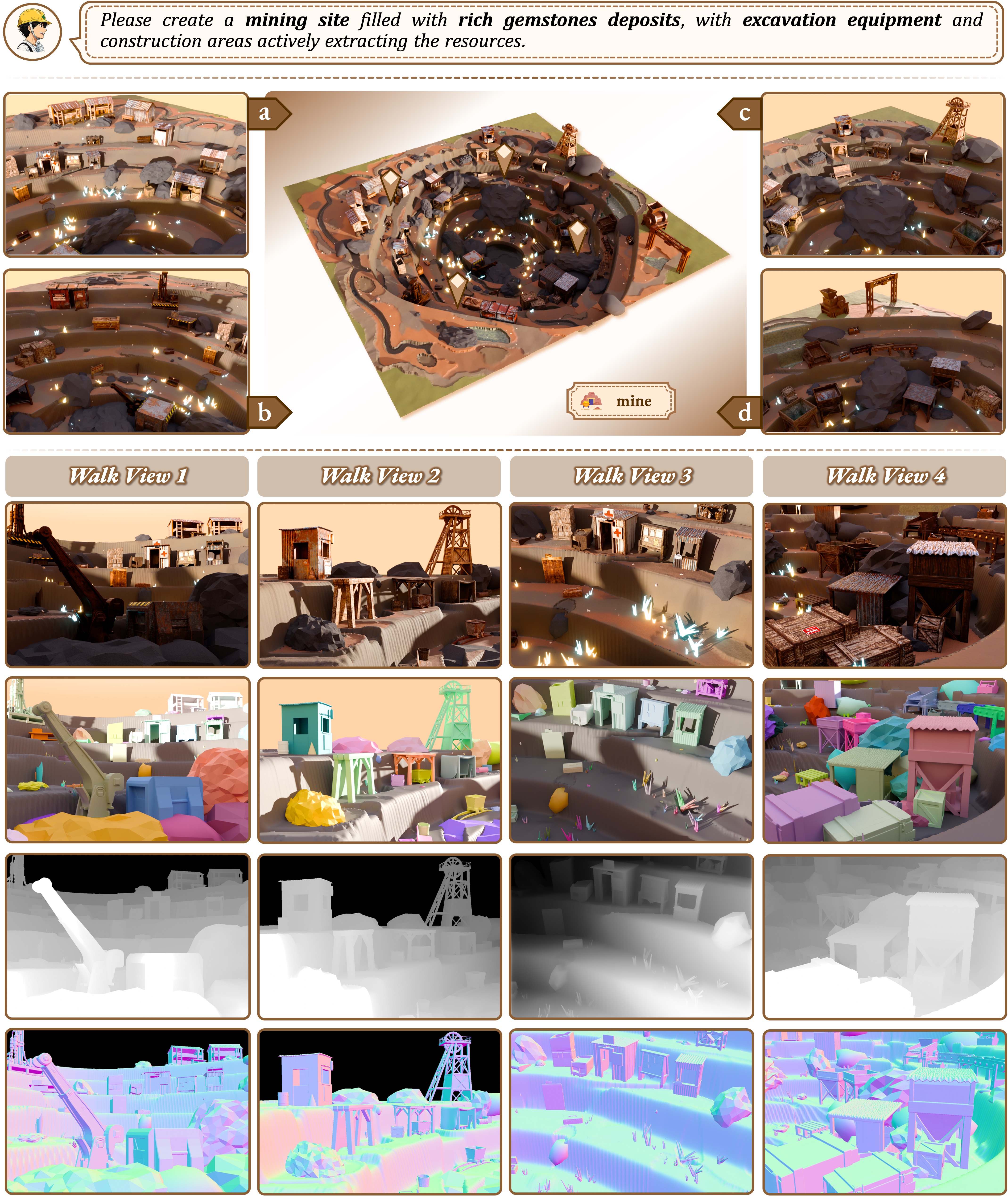}
    \caption{\textbf{Gemstone mining site.}
    The scene organizes excavation regions, exposed gemstone deposits,
    construction structures, and mining equipment within a terraced mine.
    Global, regional, and local walk views are shown with instance, depth, and
    normal renderings.}
    \label{fig:case8_mine_large}
\end{figure}

\begin{figure}[ht]
    \centering
    \includegraphics[width=1\linewidth]{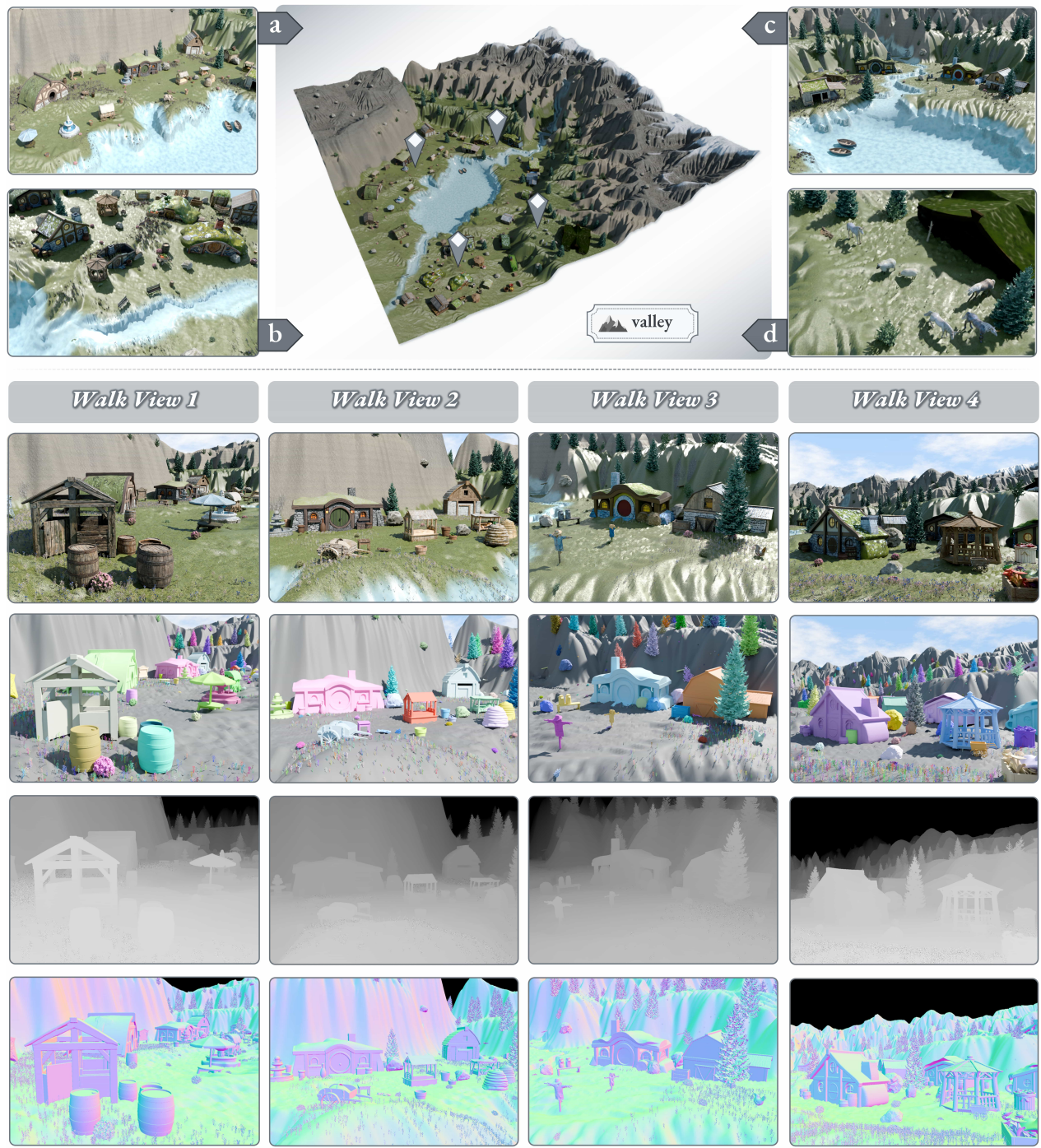}
    \caption{\textbf{Mountain valley with Hobbit-style villages.}
    Scattered settlements are embedded within a valley surrounded by steep
    hills, vegetation, and water. The figure shows the global composition,
    regional close-ups, local walk views, and their instance, depth, and
    normal renderings.}
    \label{fig:case9_valley_large}
\end{figure}

Figures~\ref{fig:case0_plain_small}--\ref{fig:case9_valley_large} present
additional scenes generated from diverse prompts. These examples
include a medieval-style village, snow-covered riverside village, a desert adventure camp surrounded
by dragons, an island with Japanese-style towns, a volcanic demon lair, a
gemstone mining site, and a mountain valley containing Hobbit-style villages.
These results show that WorldClaw can accommodate
markedly different world layouts, thematic requirements, and levels of local
complexity, underscoring its broad applicability to open-ended world creation. As in the main text, each example is shown through a global view, regional views, and local walk views, accompanied by instance, depth, and normal renderings that expose the underlying explicit scene representation.

\section{Contributors}
\label{sec:contributors}

\noindent Project Leaders: \textbf{Chunchao Guo},  \textbf{Yang Li}

\noindent Local Scene Generation:  \textbf{Jinpeng Li}, \textbf{Yang Li}, \textbf{Zilong Huang}

\noindent Global Terrain Generation: \textbf{Zilong Huang}, \textbf{Yang Li}, \textbf{Jinpeng Li}

\newpage

\bibliographystyle{plainnat}
\bibliography{reference}

\clearpage

\end{document}